\documentclass[10pt,twocolumn,letterpaper]{article}

\usepackage[pagenumbers]{cvpr} 

\usepackage{graphicx}
\usepackage{amsmath}
\usepackage{amssymb}
\usepackage{booktabs}
\usepackage{multicol}
\usepackage{subcaption}
\usepackage{amsfonts}       
\usepackage{microtype}      
\usepackage{algorithmic}
\usepackage[lined,boxed,ruled]{algorithm2e}
\usepackage{mathtools}
\usepackage{comment}
\usepackage{url}
\usepackage{color}

\usepackage[pagebackref,breaklinks,colorlinks]{hyperref}

\usepackage[capitalize]{cleveref}

\newlength{\twosubht}
\newsavebox{\twosubbox}

\newlength{\totalimgwidth}
\newlength{\imgspacingwidth}

\crefname{section}{Sec.}{Secs.}
\Crefname{section}{Section}{Sections}
\Crefname{table}{Table}{Tables}
\crefname{table}{Tab.}{Tabs.}

\newcommand{\parencite}[1]{\cite{#1}}%
\newcommand{\textcite}[1]{\cite{#1}}%

\newcommand{\citep}[1]{\cite{#1}}%
\newcommand{\citet}[1]{\cite{#1}}%

\newcommand{\todo}[1]{\textcolor{red}{[\textbf{TODO:} #1]}}

\newcommand{\bc}[1]{\textcolor{blue}{[\textbf{BC:} #1]}}

\newcommand{\blue}[1]{\textcolor{blue}{ #1}}

\newcommand{\cut}[1]{}
\newcommand{\para}[1]{{\noindent\textbf{#1}}}
\newcommand{\parai}[1]{{\noindent\textit{#1}}}

\newcommand{\LTR}{\texttt{Learn-to-Race}}
\newcommand{\ltr}{\texttt{L2R}}
\newcommand{\ourmethod}{SAGE}

\def\cvprPaperID{11664} 
\def\confName{CVPR}
\def\confYear{2022}

\begin{document}

\title{Safe Autonomous Racing via Approximate Reachability on Ego-vision}

\author{Bingqing Chen$^{1}$ ~ Jonathan Francis$^{1,2}$ ~ Jean Oh$^{1}$ ~ Eric Nyberg$^{1}$ ~ Sylvia L. Herbert$^{3}$\\
$^{1}$Carnegie Mellon University ~ 
$^{2}$Bosch Research Pittsburgh ~ 
$^{3}$University of California, San Diego \\
{\tt\small \{bingqinc, jmf1, jeanoh, ehn\}@cs.cmu.edu, sherbert@ucsd.edu}
}
\maketitle

\begin{abstract}


%



Racing demands each vehicle to drive at its physical limits, when any safety infraction could lead to catastrophic failure. In this work, we study the problem of safe reinforcement learning (RL) for autonomous racing, using the vehicle's ego-camera view and speed as input.
Given the nature of the task, autonomous agents need to be able to 1) identify and avoid unsafe scenarios under the complex vehicle dynamics, and 2) make sub-second decision in a fast-changing environment. To satisfy these criteria, we propose to incorporate Hamilton-Jacobi (HJ) reachability theory, a safety verification method for general non-linear systems, into the constrained Markov decision process (CMDP) framework. HJ reachability not only provides a control-theoretic approach to learn about safety, but also enables low-latency safety verification.  
Though HJ reachability is traditionally not scalable to high-dimensional systems, we demonstrate that with neural approximation, the HJ safety value can be learned directly on vision context---the highest-dimensional problem studied via the method, to-date.
We evaluate our method on several benchmark tasks, including Safety Gym and \LTR~(\ltr), a recently-released high-fidelity autonomous racing environment. Our approach has significantly fewer constraint violations in comparison to other constrained RL baselines in Safety Gym, and achieves the new state-of-the-art results on the \ltr~benchmark task. We release our code in the supplementary material and provide additional visualization of agent behavior at the following \href{https://sites.google.com/view/safeautonomousracing/home}{\underline{anonymized paper website}}.



\end{abstract}

\section{Introduction}
\label{sec:introduction}
Racing requires each vehicle to make sub-second decision in a fast changing environment and operate at its physical limits \cite{liniger2015optimization}, when any safety infraction could lead to catastrophic failure. Thus, autonomous racing is a particularly challenging proving ground for autonomous agents to optimize performance, while adhering to safety constraints. In the reinforcement learning (RL) literature, it is common to define safety as satisfying safety specifications \cite{ray2019benchmarking} under the constrained Markov decision process (CMDP) framework \cite{altman1999constrained}, which extends the Markov decision process (MDP) by incorporating constraints on expected cumulative costs. 

Due to the low sensor cost and high information content, camera-based perception is gaining increasing popularity in autonomous vehicles \cite{strobel2020accurate}. While end-to-end autonomous driving on visual input is an extensively-researched topic for urban driving, largely thanks to the release of the CARLA simulator \cite{dosovitskiy2017carla}, it is less so for high-speed racing, which may be partly attributed to the lack of open-source, high-fidelity simulation environments. The recent release of \LTR~(\ltr) \cite{herman2021learn} changes that and lowers the barrier of entry for autonomous racing research. 

\begin{figure*}[t!]
\vspace{-0.5cm}
\sbox\twosubbox{%
  \resizebox{\dimexpr\textwidth-1em}{!}{%
    \includegraphics[height=3cm]{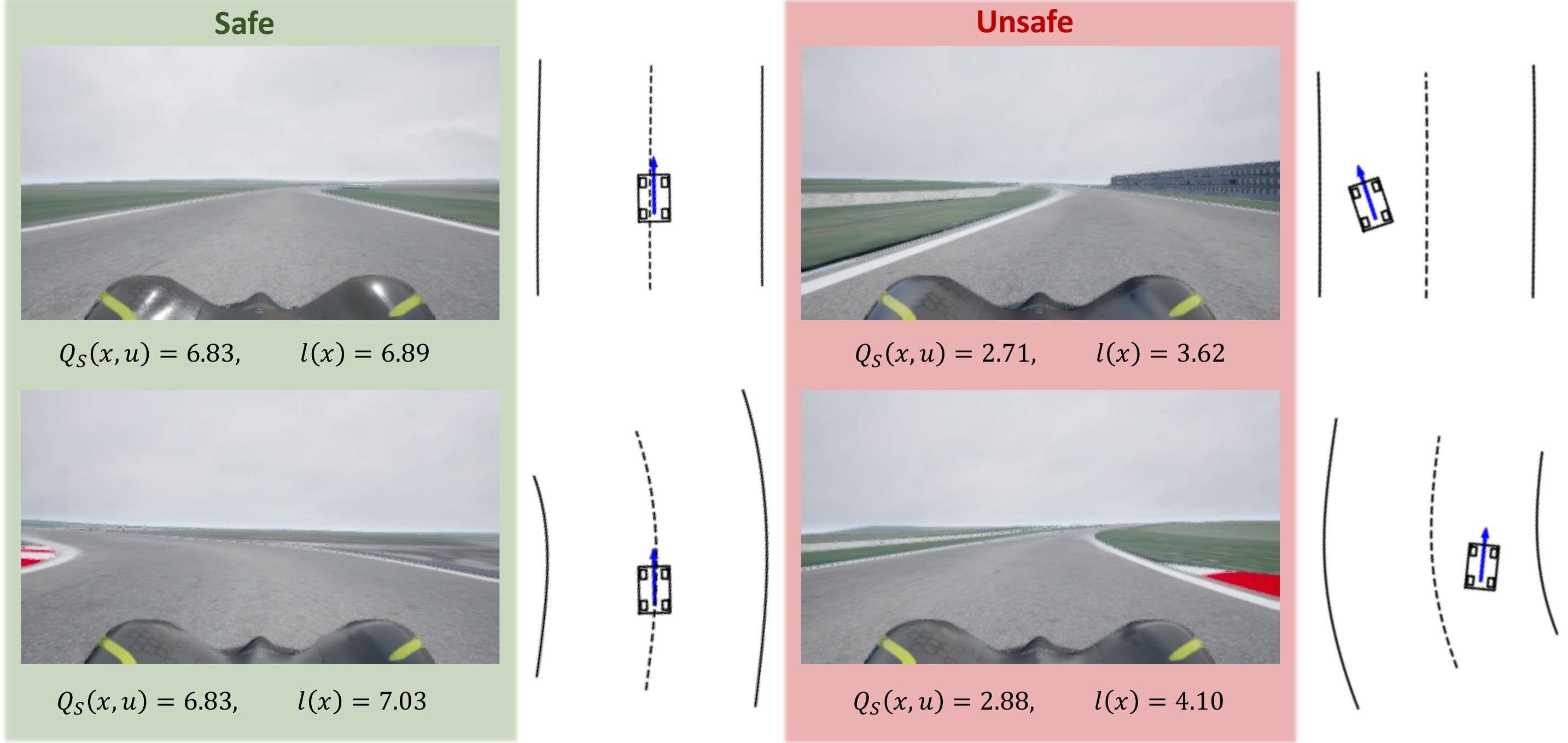}
    \hspace{0.5cm}
\includegraphics[height=3cm]{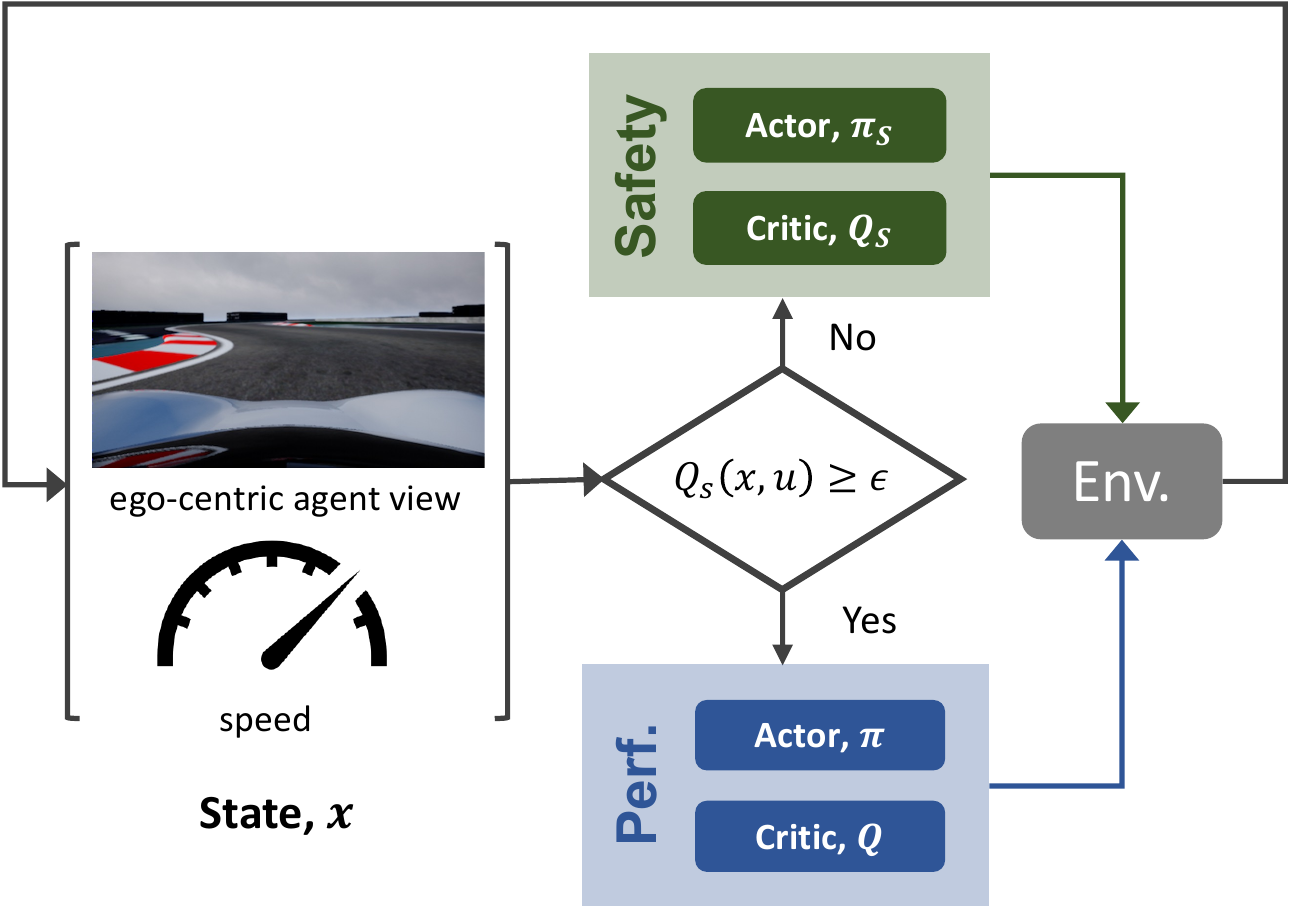}
  }
}
\setlength{\twosubht}{\ht\twosubbox}


\centering
\subcaptionbox{Examples of Safe vs. Unsafe States\footnotemark \label{fig:safe_examples}}{%
  \includegraphics[height=0.99\twosubht]{figures/fm_examples.pdf}}
  \hspace{0.5cm}
\subcaptionbox{\ourmethod~ Architecture \label{fig:spar:architecture}}{%
  \includegraphics[height=0.99\twosubht]{figures/framework_new.pdf}}
\caption[]{\ourmethod~Overview. (a) The safety critic, $Q_S$, verifies the safety of a state-action pair by checking if $Q_S(x, u)\geq\epsilon$. Some examples of safe vs. unsafe states are provided, using safety margin $\epsilon=3$, $u=\mathbf{0}$, and speed = 10m/s. (While vehicle pose is NOT available to the safety critic, we illustrate them here for reference.) (b) \ourmethod~consists of two policies, which are in charge of safety and performance, independently. The safety controller intervenes when the current state-action pair is deemed unsafe by the safety critic. 
}
\vspace{-0.1cm}
\label{fig:framework}
\end{figure*}

In this work, we study the problem of constrained RL for autonomous racing, using the vehicle's ego-camera view and speed as input. Due to the nature of the task, the autonomous agent needs to be able to 1) identify and avoid unsafe scenarios and 2) make fast safety verification given the perception data. In Figure \ref{fig:safe_examples}, we show examples of ego-camera views and the corresponding safety value $Q_S(x, u)$, estimated by our proposed safety critic, and the distance to road boundary $l(x)$. While it is straightforward to determine whether a state is safe based on the vehicle pose, which is illustrated for reference, the distinction from ego-camera views is much more subtle. Also evident from the examples is that safety does not necessarily corresponds to distance to road boundary.
Regarding the requirement for fast decision-making, in our experiments, \ltr~operates under the setting where the simulator executes the agent's command upon receiving it, and does not wait for the agent to complete its computation. Thus, high latency can adversely impact agent performance, where, as discussed in prior art \cite{strobel2020accurate}, perception stacks in autonomous race-cars account for nearly 60\% of total latency.


Given these considerations, we propose to incorporate Hamilton-Jacobi (HJ) reachability theory, a safety verification method for general non-linear systems, into the CMDP framework. HJ reachability not only provides a control-theoretical approach to learn about safety, but also enables low-latency safety verification. As a reachable set takes into consideration all possible trajectories over a specified time horizon, safety verification via HJ reachability only requires evaluating the safety value of the current state. 
Furthermore, safety verification under HJ Reachability theory does not depend on the performance policy. Thus, we can bypass the challenges involved with solving a constrained optimization problem with a neural policy, and decompose the problem of learning under safety constraints into (a) optimizing for performance, and (b) updating the safety value function. Given this intuition, we learn two policies that independently manage safety and performance (Figure \ref{fig:spar:architecture}): the performance policy focuses exclusively on optimizing performance, while the safety critic verifies if the current state is safe and intervenes when necessary. We refer to our approach as \textbf{S}afe \textbf{A}utonomous racin\textbf{G} on \textbf{E}go-vision (SAGE).
\footnotetext{Unsafe here refers to $Q_S(x, u)<\epsilon$. }

Aside from our proposed method, our key contributions are as follows. Due to the scalability issue \cite{bansal2017hamilton}, existing works on HJ reachability exclusively study problems defined on physical states, e.g., vehicle pose, instead of high-dimensional sensory inputs, such as RGB images. We demonstrate that the HJ safety value function can be learned directly on visual context via neural approximation, the highest-dimensional problem studied by HJ safety analysis to-date, thereby expanding the applications of HJ reachability to high-dimensional systems where dynamics models may not be available. 

Secondly, we compare the HJ Bellman update rule \cite{fisac2019bridging} to alternatives for learning a safety critic \cite{srinivasan2020learning, bharadhwaj2020conservative} on two classical control benchmarks, where safe 
states are known, analytically. Given the same off-policy samples, the HJ Bellman update rule is more accurate and sample efficient. 

Finally, we evaluate our methods on \texttt{Safety Gym} \cite{ray2019benchmarking} and \LTR~(\ltr) \cite{herman2021learn}, a recently-released, high-fidelity autonomous racing environment, which challenges the agent to make safety-critical decisions in a complex and fast-changing environment. While \ourmethod~is by no means free from failure, it has significantly fewer constraint violations compared to other constrained RL baselines in \texttt{Safety Gym}. We also report new state-of-the-art results on the \ltr~benchmark task, and show that incorporating a learnable safety critic grounded in control theory boosts performance especially during the initial learning phase.

\section{Related Work}
\label{sec:related_work}
\para{Autonomous racing.} One approach for autonomous racing is via model predictive control \cite{liniger2015optimization, rosolia2017autonomous, kabzan2019learning}, which solves an optimization problem with a model of the system dynamics. Aside from the challenges in modeling the complex dynamics, a significant drawback of such approach is the dependence on extensive sensor installation for localization and state estimation \cite{cai2021vision}. Another approach is to use a modular pipeline  \cite{kabzan2019learning, strobel2020accurate}, starting from perception on raw sensory inputs, to localization and object-detection, and finally to planning and control. While this approach is most commonly used in practice, disadvantages of the approach include over-complexity and error propagation \cite{yurtsever2020survey, francis2021core}. Recently, there is a lot of interest in using RL-based approaches for autonomous racing. In \cite{fuchs2020superhuman, chisari2021learning}, RL agents were trained using low-dimensional features as inputs. In \cite{chen2015deepdriving, drews2017aggressive}, intermediate features were extracted from perception pipelines to determine control actions. In \cite{cai2021vision, weiss2020deepracing}, RL agents were trained end-to-end on visual inputs by imitating expert demonstration; in \cite{cai2021vision}, a data-driven model of the environment was further utilized to train the agent by unrolling future trajectories. 

In comparison to racing, there is significantly more literature on end-to-end autonomous driving for urban scenarios \cite{codevilla2018end,  ohn2020learning, codevilla2019exploring, chen2020learning, zhang2021learning, prakash2021multi, zhang2021end, zhang2021learning}. It is beyond our scope to cover this large research field, and we refer interested readers to survey papers, such as \cite{yurtsever2020survey, grigorescu2020survey}, for more information. While we focus on high-speed racing its unique challenges, we believe the discussion here for safety analysis on ego-vision is also relevant to urban driving.

\para{Constrained reinforcement learning.} There is growing interest in enforcing some notion of safety in RL algorithms, e.g., satisfying safety constraints, avoiding worst-case outcomes, or being robust to environmental stochasticity \cite{garcia2015comprehensive}. 
We focus on the notion of safety as satisfying constraints. CMDP \cite{altman1999constrained} is a widely-used framework for studying RL under constraints, where the agent maximizes cumulative rewards, subject to limits on cumulative costs characterizing constraint violations. Solving a CMDP problem is challenging, because the policy needs to be optimized over the set of feasible states; this requires off-policy evaluation of the constraint functions, to determine whether a policy is feasible \cite{achiam2017constrained}. As a result, safety grows with experience, but requires diverse state-action pairs, including unsafe ones \cite{srinivasan2020learning}. 
Furthermore, one needs to solve a constrained optimization problem with a non-convex neural policy. This may be implemented with techniques inspired by convex optimization, such as primal-dual updates \cite{bharadhwaj2020conservative} and projection \cite{yang2020projection}, or by upper bounding the expected cost at each policy iteration \cite{achiam2017constrained}.
Most relevant to our work is  \cite{bharadhwaj2020conservative, srinivasan2020learning, thananjeyan2021recovery}, which also uses a safety critic to verify if a state is safe; we compare our control-theoretic learning rule with theirs in Section \ref{sec:toy_exp}.%




\para{Guaranteed safe control.}
Guaranteeing the safety of general continuous nonlinear systems is challenging, but there are several approaches that have been successful. These methods typically rely on knowledge of the environment dynamics. 
Control barrier functions (CBFs) provide a measure of safety with gradients that inform the acceptable safe actions \cite{ames2019control}. 
For specific forms of dynamics, e.g., control-affine \cite{cheng2019end}, and unlimited actuation bounds, this approach can be scalable to higher-dimensional systems and can be paired with an efficient online quadratic program for computing the instantaneous control \cite{cheng2019end}. Unfortunately, finding a valid control barrier function for a general system is a nontrivial task. Lyapunov-based methods \cite{chow2018lyapunov, chow2019lyapunov} suffer from the same limitation of requiring hand-crafted functions. 


HJ reachability is a technique that uses continuous-time dynamic programming to directly compute a value function that captures the optimal safe control for a general nonlinear system \cite{bansal2017hamilton, fisac2018general}. This method can provide hard safety guarantees for systems, subject to bounded uncertainties and disturbances. There are two major drawbacks to HJ reachability. The first is that the technique suffers from the curse of dimensionality and scales exponentially with number of states in the system. Because of this, the technique can only be used directly on systems of up to 4-5 dimensions \cite{bansal2017hamilton}. When using specific dynamics formulations and/or restricted controllers, this upper limit can be extended \cite{chen2018decomposition, kousik2020bridging}. Second, because of this computational cost, the value function is typically computed offline based on assumed system dynamics and bounds on uncertainties. This can lead the safety analysis to be invalid or overly conservative. 

There are many attempts in injecting some form of control theory into RL algorithms. In comparison to works that assume specific problem structure \cite{cheng2019end, dean2019safely} or existence of a nominal model \cite{cheng2019end, bastani2021safe}, our proposed approach is applicable to general nonlinear systems and does not require a model. But, we do assume access to a distance metric defined on the state space.
Our primary inspiration is recent work by \cite{fisac2019bridging} that connects HJ reachability with RL and introduces a HJ Bellman update, which can be applied to deep Q-learning for safety analysis. This method loses hard safety guarantees due to the neural approximation, but enables scalable learning of safety value function. However, an agent trained using the method in \cite{fisac2019bridging} will focus exclusively on safety. Thus, we extend the method by formulating it within the CMDP framework, thereby enabling performance-driven learning. 
\section{Preliminaries}
\label{sec:preliminaries}
\vspace{-0.2cm}
\para{Constrained MDPs.} The problem of RL with safety constraints is often formulated as a CMDP. On top of the MDP tuple $(\mathcal{X}, \mathcal{U}, R, \mathcal{F})$, where $\mathcal{X}$ is the state space, $\mathcal{U}$ is the action space, $\mathcal{F}: \mathcal{X} \times \mathcal{U}\xrightarrow{} \mathcal{X}$ characterizes the system dynamics, and $R:\mathcal{X} \times \mathcal{U} \xrightarrow{} \mathbb{R} $ is the reward function, CMDP includes an additional set of cost functions, $\{C_1,\dots,C_m\}$, where each $C_i:\mathcal{X} \times \mathcal{U} \xrightarrow{} \mathbb{R}$ maps state-action transitions to costs characterizing constraint violations.

The objective of RL is to find a policy $\pi: \mathcal{X}\xrightarrow{} \mathcal{P}(\mathcal{U})$ that maximizes the expected cumulative rewards,  $V^\pi_R(x) =\mathbb{E}_{x_k, u_k \sim \pi}  \left[\sum_{k=0}^\infty \gamma^k R(x_k, u_k)|x_0 =x\right]$, where $\gamma\in [0, 1)$ is a temporal discount factor. Similarly, the expected cumulative costs are defined as $V^\pi_{C_i}(x) = \mathbb{E}_{x_k, u_k \sim \pi} \left[\sum_{k=0}^\infty \gamma^k C_i(x_k, u_k)|x_0 =x\right]$; 
CMDP requires the policy to be feasible by imposing a limit for the costs, i.e., $V_{C_i}(\pi)\leq \chi_i, \forall i$. Putting everything together, the RL problem in a CMDP is:
\begin{align}
\label{eq:cmdp}
    \pi^{*} = \arg\max_{\pi} \;\; V^\pi_R(x)
\quad \textrm{s.t.} \;\;\ V^{\pi}_{C_i}(x)\leq \chi_i \;\;\forall i.
\end{align}

\para{HJ Reachability.} To generate the safety constraint, one can apply HJ reachability to a general nonlinear system model, denoted as $\dot{x} = f(x,u)$. Here $x\in \mathbb{R}^n$ is the state, $u$ is the control contained within a compact set $\mathcal{U}$. $f$ is assumed uniformly continuous and bounded, and Lipschitz in x for all u. For discrete-time approximations, the time step $\Delta t >0$ is used. 
We denote all allowable states as $\mathcal{K}$, for which there exists a terminal reward $l(x)$, such that $ x\in \mathcal{K} \iff l(x)\geq 0$. An $l(x)$ that satisfies this condition is the signed distance to the boundary of $\mathcal{K}$. Taking autonomous driving as an example, $\mathcal{K}$ is the drivable area and $l(x)$ is the distance to road boundary or obstacle. This set $\mathcal{K}$ is the complement of the failure set that must be avoided. The goal of this HJ reachability problem is to compute a safety value function that maps a state to its safety value, with respect to $l(x)$, over time. This is done by capturing the minimum reward achieved over time by the system applying an optimal control policy: 
\begin{equation}
    V_S(x,T) = \sup_{u(\cdot)}\min_{t\in [0, T]} l(\xi^{u}_{x,T}(t)),
\end{equation}
where $\xi$ is the state trajectory, $T<0$ is the initial time, and $0$ is the final time.  To solve for this safety value function, a form of continuous dynamic programming is applied backwards in time, from $t = 0$ to $t = T$, using the Hamilton-Jacobi-Isaacs Variational Inequality (HJI-VI):
\begin{multline}
    \min \left\{ \frac{\partial V_S}{\partial t}+ \max_{u \in \mathcal{U}}  \langle f(x, u), \nabla V_S(x) \rangle, l(x)-V_S(x, t) \right\}=0,
     \\V_S(x, 0) = l(x).
    \label{eq:HJIVI}
\end{multline}
The super-zero level set of this function is called the reachable tube, and describes all states from which the system can remain outside of the failure set for the time horizon. For the infinite-time, if the limit exists, we define the converged value function as  $V_S(x) = \lim_{T  \rightarrow -\infty}V_S(x,T)$. While the HJI-VI is difficult to solve, once solved, safety verification only requires evaluating the safety value of the current state. 

Once the safety value function is computed, the optimal safe control can be found online by solving the Hamiltonian: $\pi_S^*(x) = \arg \max_{u \in \mathcal{U}}  \langle f(x, u), \nabla V_S(x) \rangle$. This safe control is typically applied in a least-restrictive way, wherein the safety controller becomes active only when the system approaches the boundary of the reachable tube, i.e., $u\sim \pi$ if $V_S(x,T)\geq 0$ and $\pi_S^*$ otherwise. 

The newly introduced discounted safety Bellman equation \cite{fisac2019bridging} modifies the HJI-VI in \eqref{eq:HJIVI} in a time-discounted formulation for discrete time:
\begin{multline}
    V_S(x) = (1-\gamma)l(x) + \gamma \min\left\{ l(x), \max_{u\in \mathcal{U}} V_S(x+f(x,u)\Delta t) \right\},\\
    V_S(x, 0) = l(x). 
\end{multline}
This formulation induces a contraction mapping, which enables convergence of the value function when applied to dynamic programming schemes, commonly used in RL.  

\section{Safe Autonomous racinG on Ego-vision}
\label{sec:approach}


In this section, we describe our framework for safety-aware autonomous racing. We are inspired by guaranteed-safe methods, such as HJ reachability, which provides a systematic way to verify safety. Thus, we formulate our problem as a combination of constrained RL and HJ reachability theory, adopting a control-theoretic approach to learn safety. 
Building upon prior work on neural approximation of HJ Reachability \cite{fisac2019bridging}, we demonstrate that it is possible to directly update the safety value function on high-dimensional sensory input, thereby expanding the scope of applications to problems previously inaccessible. We highlight the notable aspects of our framework:
\parai{i) HJ reachability provides a control-theoretic and low-latency way to verify safety.} By incorporating HJ Reachability theory in the CMDP framework, we have a control-theoretic update rule to learn about safety and can verify safety by evaluating the safety value of the current state. Another positive outcome of the formulation is that the original constrained problem is decomposed into two unconstrained optimization problems, making our formulation more amenable to gradient-based learning.



\parai{ii) Scales to high-dimensional visual context.} Compared to standard HJ Reachability methods, whose computational complexity scales exponentially with the state dimension, we updated the safety value directly on vision embedding, with neural approximation. This is the highest-dimensional problem studied studied via HJ reachability to-date.  
\begin{algorithm}[t!] 
\SetAlgoLined
\textbf{Initialize: } Performance  actor $\pi$ and critic $Q$; \\
\textbf{Initialize: } Safety actor $\pi_S$ and critic $Q_S$;  \\
 \For{i = 0, $\dots$, \# Episodes} 
{$x $ = env.reset()\\
 \While{not terminal}{
 {u $\sim \pi(x)$;\\
 \emph{// The safe actor intervenes when the current state-action is deemed unsafe by the safety critic.}\\
  \If{$Q_{S}(x, u)< \epsilon$}{
   $u \sim \pi_{S}(x)$
  }
 $x', r $ = env.step($u$)\\
 \emph{// Update performance actor-critic and safety actor-critic. See Appendix \ref{sec:algorithm} for details.}\\
 }
}
 }
 \caption{\ourmethod-Environment Interaction}
 \label{alg:spar}

\end{algorithm}
 
\begin{figure*}[t!]
\vspace{-1cm}
\sbox\twosubbox{%
  \resizebox{\dimexpr\textwidth-1em}{!}{%
    \includegraphics[height=3cm]{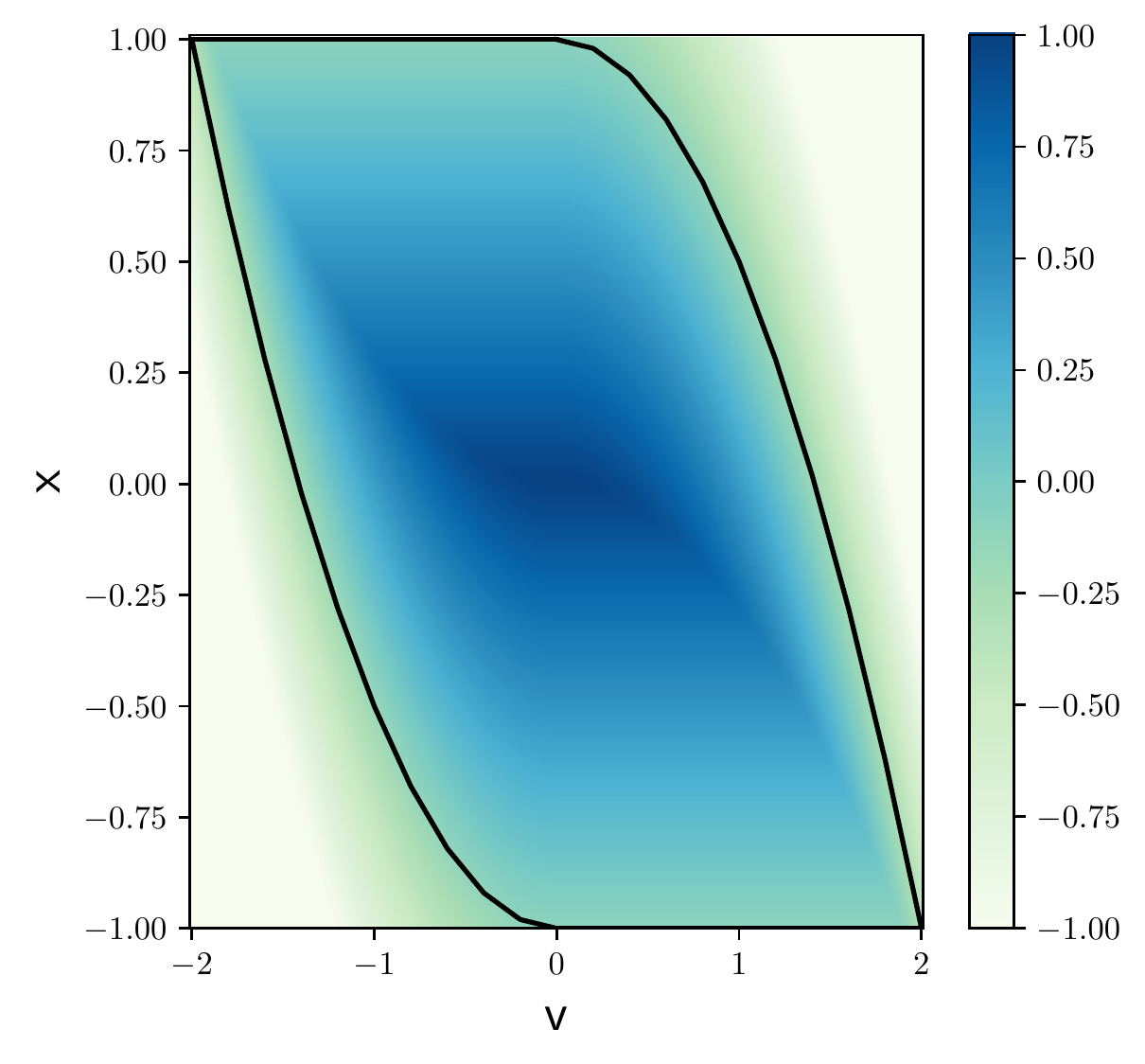}
\includegraphics[height=3cm]{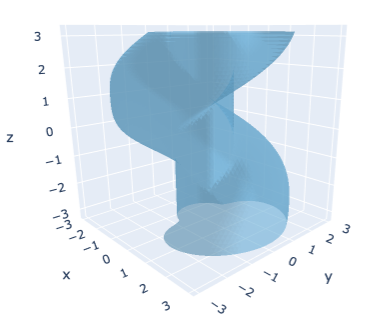}
\includegraphics[height=3cm]{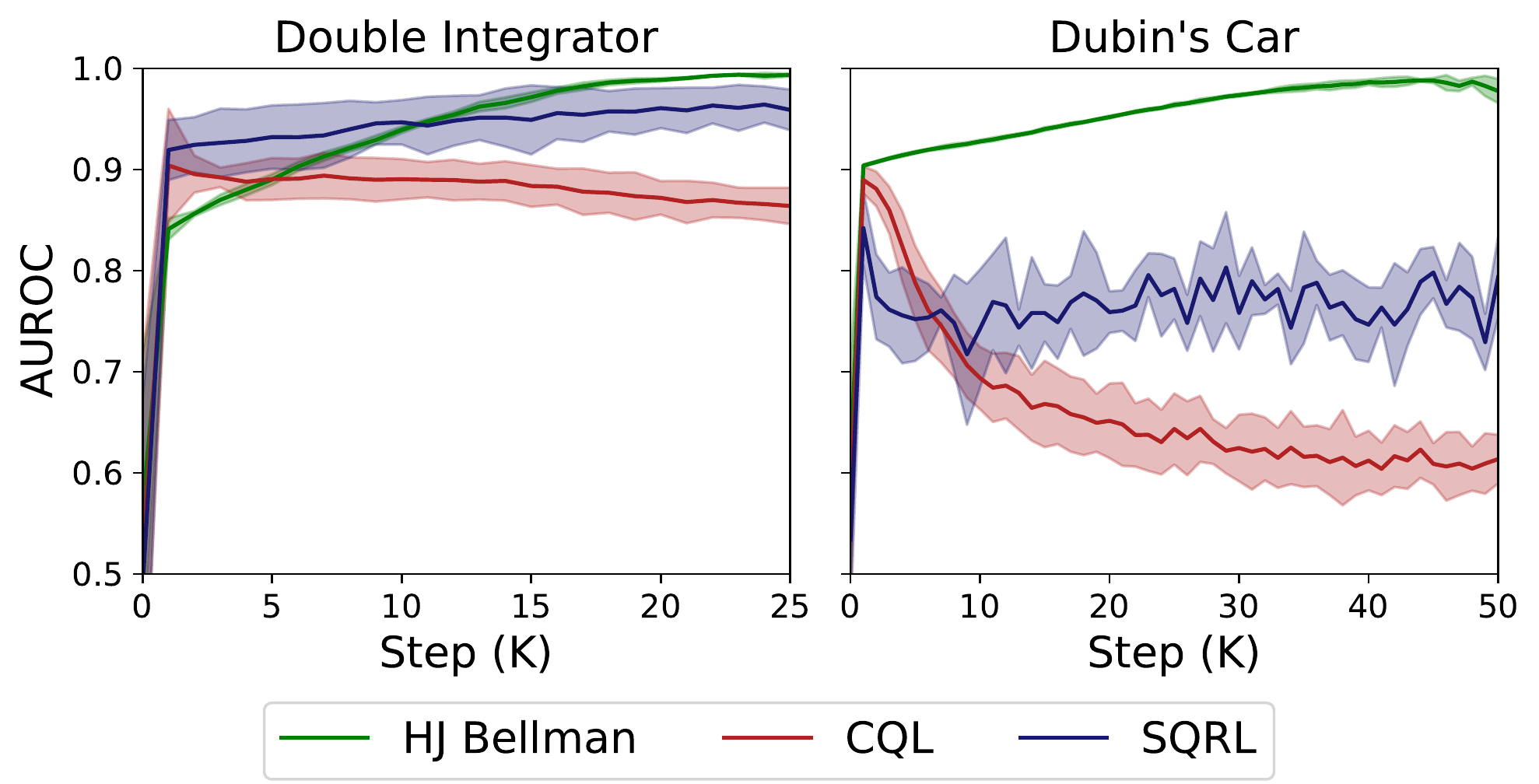}
  }
}
\setlength{\twosubht}{\ht\twosubbox}


\centering
\subcaptionbox{Double Integrator\label{fig:DoubleIntegrator}}{%
  \includegraphics[height=0.9\twosubht]{figures/DoubleIntegrator.pdf}}
\subcaptionbox{Dubins' Car\label{fig:DubinsCar}}{%
  \includegraphics[height=0.95\twosubht]{figures/DubinsCar.png}}
\subcaptionbox{Performance comparison of learning rules (averaged over 5 random seeds)\label{fig:toy_accuracy}}{%
  \includegraphics[height=\twosubht]{figures/CompALL.pdf}
  }
\caption{We use two classical control benchmarks, \textit{double integrator} and \textit{Dubins' car}, to evaluate the performance of different learning rules for safety analysis. (a) shows the safety value function of the double integrator and the black line delineates $V_S(x)=0$, within which the particle can remain within the allowable range of $x\in [-1, 1]$. (b) shows the iso-surface of the safety value function at 0, i.e., $V_S(x)=0$, for Dubins' Car, within which the car can reach a unit circle at the origin. The performance comparison is summarized in (c).
}
\vspace{-0.5cm}
\label{fig:classical_control_bench}
\end{figure*}

\para{Problem formulation.}  We inject HJ reachability theory into the CMDP framework. Starting with Eqn. \ref{eq:cmdp}, we can interpret the negative of a cost as a reward for safety and, without loss of generality, reverse the direction of the inequality constraint. Recall that the super-zero level set of the safety value function, i.e., $\{x|V_s(x)\geq0\}$, designates all states from which the system can remain within the set of allowable states, $\mathcal{K}$, over infinite time horizon.
Thus, the safety value function derived from HJ Reachability can be plugged into CMDP (Eqn. \ref{eq:problem_formulation}): %
\begin{equation}
\label{eq:problem_formulation}
\pi^{*} = \arg\max_{\pi} \;\; V^{\pi}_R(x),
\quad \textrm{s.t.} \;\;\ V_{\text{\textit{S}}}(x) \geq \epsilon,
\end{equation}
where $\epsilon\geq0$ is introduced as a safety margin. A key difference from the original CMDP formulation (Eqn. \ref{eq:cmdp}) is that constraint satisfaction, $V_S(x)\geq \epsilon$, no longer depends on the policy, $\pi$. Thus, we can bypass the challenges of solving CMDPs (Section \ref{sec:related_work}) and decompose learning under safety constraints into optimizing for performance and updating safety value estimation. While a number of works have similar dual-policy architecture \cite{cheng2019end, bastani2021safe, thananjeyan2021recovery}, ours design is informed by HJ Reachability theory. Another difference is that HJ Reachability considers safety as absolute, and there is no mechanism to allow for some level of safety infraction, and thus $\chi_i$ in Eqn. \ref{eq:cmdp} is not longer present.  




\para{Update of Safety Critic.} We apply HJ Bellman update, in place of standard Bellman backup, to learn the safety value function. The learning rule proposed by \cite{fisac2019bridging} is defined on discrete action space, which we modify for continuous action space (Eqn. \ref{eqn:safey_update}). While the safety actor is sub-optimal during learning, the resulting HJ Bellman target is an under-estimation of the safety value, as $Q_S(x', u')\leq \max_{u'\in \mathcal{U}} Q_S(x', u')$. Note that $Q_S(x, u)$ is updated model-free using state-action transitions, and only additionally requires $l(x)$. We assume $l(x)$ can be acquired from the vehicle's sensing capability \cite{bmw_2021} or estimated from perception \cite{chen2015deepdriving}.
\begin{equation}
\label{eqn:safey_update}
\begin{split}
    Q_S(x, u) &= (1-\gamma)l(x) + \gamma\min  \{l(x), Q_S(x', u')\},\\
    u'&\sim \pi_S(x').
\end{split}
\end{equation}

\para{\ourmethod.}  We propose \textsc{\ourmethod},  which consists of a performance policy and a safety policy. The safety backup controller is applied in a least restrictive way, only intervening when the RL agent is about to enter into an unsafe state, i.e., $u\sim \pi$, if $Q_S(x, u)\geq \epsilon$ and $u\sim \pi_S$ otherwise. The performance policy may be implemented with any RL algorithm. Since we expect the majority of samples to be from the performance policy, it is more appropriate to update the safety actor critic with an off-policy algorithm. In this work, we base our implementation of the safety actor critic on soft-actor critic (SAC) \parencite{haarnoja2018soft}. The safety critic is updated with Eqn. \ref{eqn:safey_update}, and the safety actor is updated via policy gradient through the safety critic, i.e., $\nabla_u Q_S(x, u)$.  Algorithm \ref{alg:spar} provides an overview for \texttt{\ourmethod} and a detailed version is presented in Appendix \ref{sec:algorithm}.


\section{Experiments}
\label{sec:experiments}
We evaluate \texttt{\ourmethod} on three sets of benchmarks, of increasing difficulty. While the our intended application is autonomous racing, the first two set of benchmarks can be considered as some abstraction of vehicles with the objective of avoiding obstacles and/or moving towards goals. Firstly, we evaluate on two classical control tasks, where the safe vs. unsafe states are known analytically, and we compare the HJ Bellman update used in \texttt{\ourmethod} to alternatives for learning safety critics in the literature. Secondly, we compare \texttt{\ourmethod} to constrained RL baselines in Safety Gym. Finally, we challenge \texttt{\ourmethod} in \LTR~and conduct ablation to better understand how different components of \texttt{\ourmethod} contribute to its performance. 
\subsection{Experiment: Classical Control Benchmarks}\label{sec:toy_exp}

\begin{figure*}[!]
    \centering
    \includegraphics[width=0.8\linewidth]{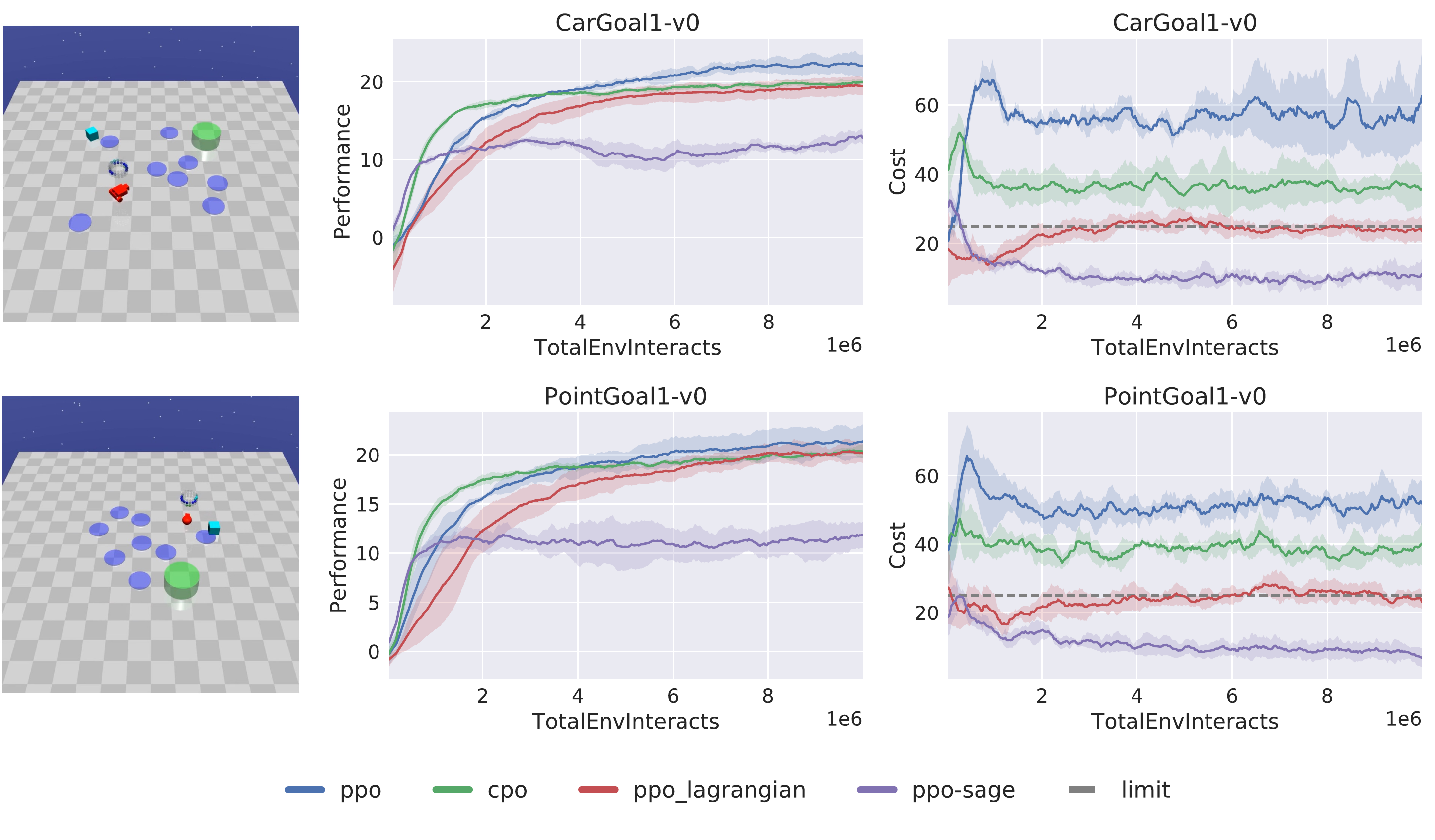}
    \caption{Performance of \ourmethod~with comparison to baselines in the CarGoal1-v0 (top row) and PointGoal1-v0 (bottom row) benchmarks (averaged over 5 random seeds). In Goal tasks, agents must navigate to observed goal locations (indicated by the green regions), while avoiding obstacles (e.g., vases in cyan, and hazards in blue).}
    \label{fig:gym_results}
\end{figure*}
As mentioned earlier, safety critics have been trained in other works \cite{bharadhwaj2020conservative, srinivasan2020learning} with different learning rules. The objective here is to compare the HJ Bellman update with alternatives. Thus, we focus on safety analysis with off-policy samples, and 
evaluate on two classical control benchmarks \textit{Double Integrator} \cite{fisac2019bridging} and \textit{Dubins' Car} \cite{bansal2017hamilton}, where the safe / lively\footnote{Liveness refers the ability to reach the specified goal \cite{hsu2021safety}.} states (Figure \ref{fig:DoubleIntegrator} and \ref{fig:DubinsCar}) and the optimal safety controller are known analytically. Double Integrator characterizes a particle moving on the x-axis, with velocity $v$. By controlling the acceleration, the objective is to keep the particle on a bounded range on x-axis. Dubins' Car is a simplified car model, where the car moves at a constant speed. By controlling the turning rate, the goal is to reach a unit circle regardless of the heading. More information on the two tasks are provided in Appendix \ref{sec:toy_model}.

In this experiment, we generate state-action pairs with a random policy, and evaluate the safety value function with respect to the optimal safety controller, $\pi^*_S$. In both \textit{Safety Q-functions for RL} (SQRL) \cite{srinivasan2020learning} and the \textit{Conservative Safety Critic} (CSC) \cite{bharadhwaj2020conservative}, the safety value function is defined as the expected cumulative cost, i.e., $Q^\pi_C(x, u) \coloneqq \mathbb{E}_{x_k, u_k \sim \pi} \left[\sum_{k=0}^\infty \gamma^k C(x_k)|x_0=x, u_0=u\right]$, where $C(x_k)=1$ if a failure occurs at $x_k$ and 0 otherwise. In this case, both the environment and optimal safety policy are deterministic. Thus, by definition, $Q^{\pi_S^*}_C(x, \pi_S^*(x))$ should be 0 if $x$ is a safe state. SQRL uses the standard Bellman backup to propagate the failure signal. On top of that, CSC uses conservative Q-learning (CQL) \cite{kumar2020conservative} to correct for difference between the behavior policy, i.e., the random policy, and the evaluation policy, i.e., the optimal safety policy, and overestimates $Q_C$ to err on the side of caution.

Since the safe vs. unsafe states are known for these benchmark tasks, we can directly compare the performance of these safety critics learned with different learning rules (Figure \ref{fig:toy_accuracy}). While the theoretical cut-off for safe vs. unsafe states is 0, the performance of SQRL is sensitive to the choice of the cut-off; thus, we report AUROC instead. For both CQL and SQRL, we do a grid search around the hyperparameters used in the original papers and report the best results. The implementation details and additional results are included in Appendix \ref{sec:toy_comp}. Directly applying Bellman update for safety analysis, as in SQRL, performs reasonably well on \textit{Double Integrator}, but not on the more challenging \textit{Dubins' Car}. In our experiment, CQL consistently under-performs SQRL. In comparison, HJ Bellman update has AUROC close to 1 on both tasks and has very small variance over different runs. It is worth-noting that the result with the HJ Bellman update is achieved without explicitly addressing the distribution mismatch \cite{voloshin2019empirical}, which typically challenges off-policy evaluation problems.
This experiment only compares the efficacy of the different learning rules for safety critic given the same off-policy samples, and does not intend to compare other aspects of SQRL and CSC.

One caveat is that SQRL and CQL uses a binary signal for failures, while HJ Bellman update has access to the distance, $l(x)$. On one hand, HJ Bellman update does assume more information. On the other hand, it may be more practical to learn safety from distance measurements then experiencing failures. Applied to autonomous driving, this translates to learning to avoid obstacle from distance measurements that are becoming prevalent on cars with assisted driving capabilities \cite{bmw_2021}, in comparison to experiencing collisions. 


\begin{figure*}[t]
\sbox\twosubbox{%
  \resizebox{\dimexpr\textwidth-1em}{!}{%
    \includegraphics[height=3cm]{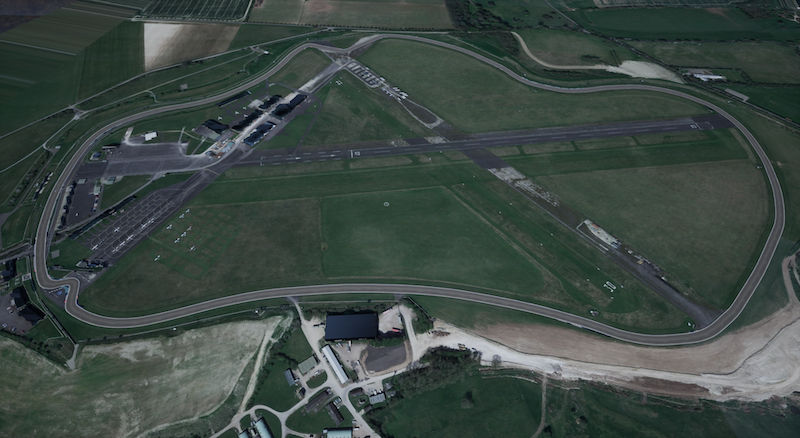}
\includegraphics[height=3cm]{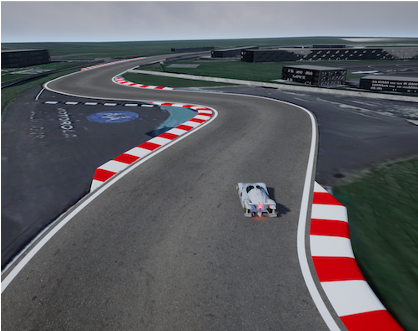}
\includegraphics[height=3cm]{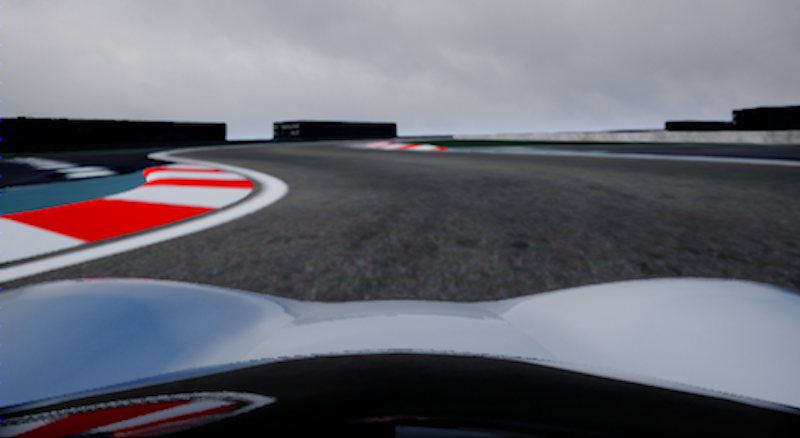}
  }
}
\setlength{\twosubht}{\ht\twosubbox}


\centering
\subcaptionbox{Aerial\label{fig:racetrack:aerial}}{%
  \includegraphics[height=0.95\twosubht]{figures/racetrack_overhead_view.png}}
\subcaptionbox{Third-person\label{fig:racetrack:third_person}}{%
  \includegraphics[height=0.95\twosubht]{figures/racetrack_speedtrap_narrow.png}}
\subcaptionbox{Ego-view\label{fig:racetrack:ego_view}}{%
  \includegraphics[height=0.95\twosubht]{figures/racetrack_agent_egoview.png}
  }
\caption{We use the \LTR~(\ltr) framework \cite{herman2021learn} for evaluation; this environment provides simulated racing tracks that are modeled after real-world counterparts, such as the famed Thruxton Circuit in the UK (\texttt{Track01:Thruxton}, (a)). Here, learning-based agents can be trained and evaluated according to challenging metrics and realistic vehicle and environmental dynamics, making \ltr~a compelling target for safe reinforcement learning. Each track features challenging components for autonomous agents, such as sharp turns (shown in (b)), where SAGE only uses ego-camera views (shown in (c)) and speed.
}
\vspace{-0.5cm}
\label{fig:racetrack}
\end{figure*}

\subsection{Experiment: \texttt{Safety Gym}} \label{exp:gym}

We additionally evaluate our proposed approach, \ourmethod, in Safety Gym \parencite{ray2019benchmarking}. Specifically, we evaluate on the standard CarGoal1-v0 and PointGoal1-v0 benchmarks, where the agent navigates to a goal while avoiding hazards. We compare \ourmethod~against baselines including: Constrained Policy optimization (CPO) \parencite{achiam2017constrained}, an unconstrained RL algorithm (Proximal Policy optimization (PPO) \parencite{schulman2017proximal}), and its Lagrangian variant (PPO-Lagrangian). By default, distance measurements from LiDAR are available to all baselines in these benchmarks, and thus \ourmethod~has direct access to $l(x)$. Episodic Performance and Cost curves are shown in Figure \ref{fig:gym_results} and  implementation details are included in Appendix \ref{app:sec:sgym_details}. 

PPO-SAGE has significantly fewer constraint violations, compared to other baselines, and the number of violations decreases over time. 
While CPO and PPO-Lagrangian take into account that a certain number of violations are permissible, there is no such mechanism in SAGE, as HJ Reachability theory defines safety in an absolute sense. While the inability to allow for some level of safety infractions, unfortunately, compromises performance, SAGE learns mature obstacle-avoidance behaviors, compared to some policies, which may ignore traps in favor of fast navigation to goal locations. Violations that do occur in SAGE result from neural approximation error, and the number of violations decreases over time as the safety actor-critic gains experience, despite the randomized and constantly-changing episodic layouts.  

\vspace{-0.2cm}

\subsection{Experiment: \LTR}
\para{Task Overview.} In this paper, we evaluate our approach using the Arrival Autonomous Racing Simulator, through the newly-introduced and OpenAI-gym compliant \LTR~(\ltr) task and evaluation framework \parencite{herman2021learn}. \ltr~provides multiple simulated racing tracks, modeled after real-world counterparts, such as Thruxton Circuit in the UK (\texttt{Track01:Thruxton}; see Figure \ref{fig:racetrack}). 
 \ltr~can provide access to RGB images from any specified location, semantic segmentation, and  vehicle states (e.g., pose, velocity). In each episode, an agent is spawned on the selected track. At each time-step, it uses its observations to determine normalized steering angle and acceleration. All learning-based agents receive the reward specified by \ltr, which is formulated as a weighted sum of reward for driving fast and penalty for leaving the drivable area; the main objective is to complete laps in as little time as possible. Additional metrics are defined to evaluate driving quality.

\para{Implementation Details.} To characterize the performance of our approach, we report results on the Average  Speed and the Episode Completion Percentage (\textit{ECP}) metrics \parencite{herman2021learn} as proxies for agent performance and safety, respectively. We report other metrics defined by \ltr~in Appendix \ref{sec:add_results}. 

We use \texttt{Track01:Thruxton} in \ltr~(Fig. \ref{fig:racetrack}) for all stages of agent interaction with the environment. During training, the agent is spawned at random locations along the race track and uses a stochastic policy. During evaluation, the agent is spawned at a fixed location and uses a deterministic policy. The episode terminates when the agent successfully finishes a lap, leaves the drivable area, collides with obstacles, or does not progress for a number of time-steps. For each agent, we report averaged results across 5 random seeds, evaluated every 5000 steps over an episode (one lap). We use SAC as the performance policy, and all agents only have access to ego-camera view (Figure \ref{fig:racetrack:ego_view}) and speed, unless specified otherwise. The implementation, including network architecture and hyperparameters, are detailed in Appendix \ref{sec:implementation_details}.





\begin{figure*}[t]
\vspace{-0.0cm}
    \centering
    \includegraphics[width=0.9\linewidth]{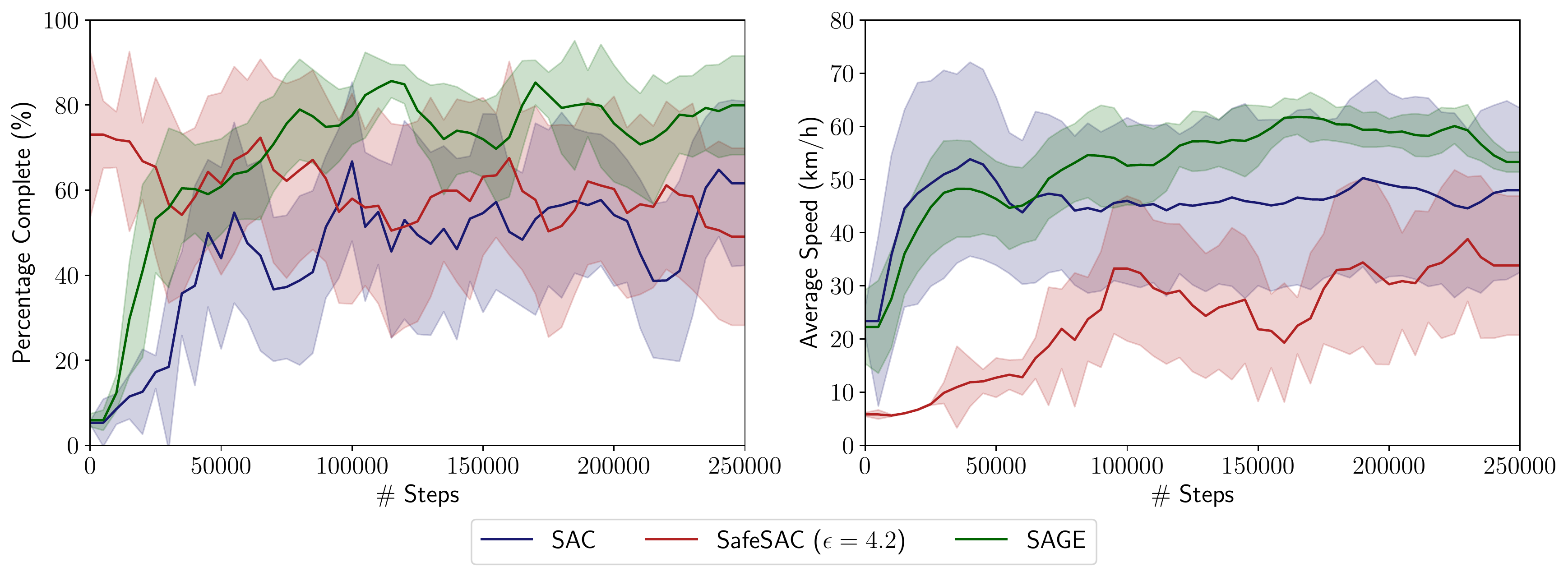}
    \caption{\textbf{Left:} Episode percent completion and \textbf{Right:} speed evaluated every 5000 steps over an episode (a single lap) and averaged over 5 random seeds. Results reported based on \texttt{Track01:Thruxton} in \ltr.}
    \label{fig:main_results}
\end{figure*}

\para{Ablation Study.} 
To demonstrate the benefit of utilizing domain knowledge in the form of a nominal model, we use a kinematic bike model \cite{kong2015kinematic} to calculate the safety value and derive the corresponding safety controller, detailed in Appendix in \ref{sec:nominal_ac}. We refer to this as the \textit{static actor-critic}. In all our experiments, only the static actor-critic has access to vehicle pose, specifically location and heading. We evaluate the performance of this static actor-critic by coupling a random agent with it (\texttt{SafeRandom}). We test \texttt{SafeRandom} on a series of safety margins to account for unmodelled dynamics; the performance averaged over 10 random seeds is summarized in Figure \ref{fig:SafeRandom}. For instance, $\epsilon\geq4.2$ achieves 80+\% ECP, in comparison to 0.5\% ECP by \texttt{Random} agent. 

We examine the effect of having a safety controller, by comparing \texttt{SAC} with an instance of itself that is coupled with a static safety actor-critic (\texttt{SafeSAC}). We set the safety margin $\epsilon$ to be 4.2, based on empirical results from \texttt{SafeRandom}. We also compare the performance of using the static safety actor-critic (\texttt{SafeSAC}) and a dynamically-updating one (\texttt{\ourmethod}). Since the \texttt{\ourmethod} agent is expected to have a better characterization of the safety value, the agent no longer depends on a large safety margin to remain safe and thus \texttt{\ourmethod} uses a safety margin of 3.0m, which accounts for the vehicle dimensions\footnote{The HJ reachable tube is computed with respect to the back axle of the vehicle and does not account for the physical dimension of the vehicle. Using the car length as the safety margin is a rough engineering estimate.}. We also report results of \texttt{SafeSAC} with the same safety margin in in Appendix \ref{sec:add_results}.

\para{Results.}
\label{sec:result}
The performance comparison between different agents is summarized in Figure \ref{fig:main_results}. 
In interpreting the results, note that a single lap in \texttt{Track01:Thruxton} is 3.8km, whereas CARLA, the \textit{de facto} environment for urban driving research, has in total 4.3km drivable roads in the original benchmark \cite{codevilla2019exploring}. Thus, successfully completing an episode, i.e., a lap, is quite challenging. 

\parai{The static safety actor-critic significantly boosts initial safety performance.} With the help of the static safety actor-critic, the \texttt{SafeSAC} can complete close to 80\% of a lap, in comparison to slightly more than 5\% with \texttt{SAC}. This, again, showcases the benefit of injecting domain knowledge in the form of a nominal model. However, there are two notable limitations with the static safety controller. Firstly, it is extremely conservative, hard-braking whenever the vehicle is less safe. As a result, the \texttt{SafeSAC} agent has an initial speed of less than 10km/h. Secondly, as the \texttt{SAC} learns to avoid activating the safety controller and drive faster, the static safety controller is no longer able to recover the vehicle from marginally safe states. In fact, by applying the `optimal' safety action from Eqn. \ref{eq:safety_control}, i.e., maximum brake and steer towards centerline, the vehicle will lose traction  and spin out of control. As a result, the ECP actually decreases over time for \texttt{SafeSAC}.

\parai{\ourmethod~learns safety directly from vision context and can recover from marginally safe states more smoothly.} Having a safety actor-critic that is dedicated to learning about safety significantly boosted the initial safety performance of \texttt{\ourmethod}, in comparison to the \texttt{SAC} agent, even though both the performance and the safety actor-critics are  randomly initialized. Moreover, this shows that the safety value function can be learned from scratch on vision embedding. In practice, we envision the safety actor-critic to be warm-started with the nominal model or observational data, and fine-tuned by interactions with the environment. Furthermore, the learnable safety actor-critic can recover from marginally safe states smoothly, avoiding the two undesirable behaviors from the static actor-critic. A qualitative comparison of such behaviors is available at the \href{https://sites.google.com/view/safeautonomousracing/home}{\underline{anonymized paper website}}. 
 While \texttt{\ourmethod} outperforms other baselines, there is still significant performance gap with human, as the speed record at Thruxton Circuit is 237 km/h (average speed).




\section{Conclusion}
\label{sec:conclusion}

In this paper, we propose \texttt{SAGE} for end-to-end autonomous racing, which can learn to identify unsafe states from ego-camera views and recover from unsafe states, despite the complex dynamics with unstable regimes. We demonstrate on two classical control benchmarks that the HJ Bellman update is more effective than alternatives for learning the safety critic. Compared to constrained RL baselines in the Safety Gym, we show that \texttt{\ourmethod} has significantly fewer constraint violations. We report the new state-of-the-art result on \LTR, and we demonstrate that the safety value can be learned directly on visual context, thereby expanding HJ reachability to broader applications. 

Throughout our experiments, we find it is highly effective to inject domain knowledge, in the form of nominal model or control-theoretic learning rule. In our experiments, the safety actor-critics were randomly initialized. But, in practice, we expect it to be pre-trained with a nominal model and/or observational data, prior to interacting with the environment. While neural approximation enables us to scale HJ reachability to high-dimensional visual inputs, we unfortunately lose the hard guarantees on safety. An important next-step is to characterize neural approximation error and find ways to retain the safety guarantees with function approximators.

\clearpage

{\small
\bibliographystyle{ieee_fullname}
\bibliography{egbib}

\begin{thebibliography}{10}\itemsep=-1pt

\bibitem{SpinningUp2018}
Joshua Achiam.
\newblock {Spinning Up in Deep Reinforcement Learning}.
\newblock 2018.

\bibitem{achiam2017constrained}
Joshua Achiam, David Held, Aviv Tamar, and Pieter Abbeel.
\newblock Constrained policy optimization.
\newblock In {\em International Conference on Machine Learning}, pages 22--31.
  PMLR, 2017.

\bibitem{altman1999constrained}
Eitan Altman.
\newblock {\em Constrained Markov decision processes}, volume~7.
\newblock CRC Press, 1999.

\bibitem{ames2019control}
Aaron~D Ames, Samuel Coogan, Magnus Egerstedt, Gennaro Notomista, Koushil
  Sreenath, and Paulo Tabuada.
\newblock Control barrier functions: Theory and applications.
\newblock In {\em 2019 18th European Control Conference (ECC)}, pages
  3420--3431. IEEE, 2019.

\bibitem{bansal2017hamilton}
Somil Bansal, Mo Chen, Sylvia Herbert, and Claire~J Tomlin.
\newblock Hamilton-jacobi reachability: A brief overview and recent advances.
\newblock In {\em 2017 IEEE 56th Annual Conference on Decision and Control
  (CDC)}, pages 2242--2253. IEEE, 2017.

\bibitem{bastani2021safe}
Osbert Bastani.
\newblock Safe reinforcement learning with nonlinear dynamics via model
  predictive shielding.
\newblock In {\em 2021 American Control Conference (ACC)}, pages 3488--3494.
  IEEE, 2021.

\bibitem{bharadhwaj2020conservative}
Homanga Bharadhwaj, Aviral Kumar, Nicholas Rhinehart, Sergey Levine, Florian
  Shkurti, and Animesh Garg.
\newblock Conservative safety critics for exploration.
\newblock {\em arXiv preprint arXiv:2010.14497}, 2020.

\bibitem{bmw_2021}
{BMW}.
\newblock Automotive sensors – the sense organs of driver assistance systems,
  Sep 2021.

\bibitem{cai2021vision}
Peide Cai, Hengli Wang, Huaiyang Huang, Yuxuan Liu, and Ming Liu.
\newblock Vision-based autonomous car racing using deep imitative reinforcement
  learning.
\newblock {\em IEEE Robotics and Automation Letters}, 6(4):7262--7269, 2021.

\bibitem{chen2015deepdriving}
Chenyi Chen, Ari Seff, Alain Kornhauser, and Jianxiong Xiao.
\newblock Deepdriving: Learning affordance for direct perception in autonomous
  driving.
\newblock In {\em Proceedings of the IEEE international conference on computer
  vision}, pages 2722--2730, 2015.

\bibitem{chen2020learning}
Dian Chen, Brady Zhou, Vladlen Koltun, and Philipp Kr{\"a}henb{\"u}hl.
\newblock Learning by cheating.
\newblock In {\em Conference on Robot Learning}, pages 66--75. PMLR, 2020.

\bibitem{chen2018decomposition}
Mo Chen, Sylvia~L Herbert, Mahesh~S Vashishtha, Somil Bansal, and Claire~J
  Tomlin.
\newblock Decomposition of reachable sets and tubes for a class of nonlinear
  systems.
\newblock {\em IEEE Transactions on Automatic Control}, 63(11):3675--3688,
  2018.

\bibitem{cheng2019end}
Richard Cheng, G{\'a}bor Orosz, Richard~M Murray, and Joel~W Burdick.
\newblock End-to-end safe reinforcement learning through barrier functions for
  safety-critical continuous control tasks.
\newblock In {\em Proceedings of the AAAI Conference on Artificial
  Intelligence}, volume~33, pages 3387--3395, 2019.

\bibitem{chisari2021learning}
Eugenio Chisari, Alexander Liniger, Alisa Rupenyan, Luc Van~Gool, and John
  Lygeros.
\newblock Learning from simulation, racing in reality.
\newblock In {\em 2021 IEEE International Conference on Robotics and Automation
  (ICRA)}, pages 8046--8052. IEEE, 2021.

\bibitem{chow2018lyapunov}
Yinlam Chow, Ofir Nachum, Edgar Duenez-Guzman, and Mohammad Ghavamzadeh.
\newblock A lyapunov-based approach to safe reinforcement learning.
\newblock {\em arXiv preprint arXiv:1805.07708}, 2018.

\bibitem{chow2019lyapunov}
Yinlam Chow, Ofir Nachum, Aleksandra Faust, Edgar Duenez-Guzman, and Mohammad
  Ghavamzadeh.
\newblock Lyapunov-based safe policy optimization for continuous control.
\newblock {\em arXiv preprint arXiv:1901.10031}, 2019.

\bibitem{codevilla2018end}
Felipe Codevilla, Matthias M{\"u}ller, Antonio L{\'o}pez, Vladlen Koltun, and
  Alexey Dosovitskiy.
\newblock End-to-end driving via conditional imitation learning.
\newblock In {\em 2018 IEEE International Conference on Robotics and Automation
  (ICRA)}, pages 4693--4700. IEEE, 2018.

\bibitem{codevilla2019exploring}
Felipe Codevilla, Eder Santana, Antonio~M L{\'o}pez, and Adrien Gaidon.
\newblock Exploring the limitations of behavior cloning for autonomous driving.
\newblock In {\em Proceedings of the IEEE/CVF International Conference on
  Computer Vision}, pages 9329--9338, 2019.

\bibitem{dean2019safely}
Sarah Dean, Stephen Tu, Nikolai Matni, and Benjamin Recht.
\newblock Safely learning to control the constrained linear quadratic
  regulator.
\newblock In {\em 2019 American Control Conference (ACC)}, pages 5582--5588.
  IEEE, 2019.

\bibitem{dosovitskiy2017carla}
Alexey Dosovitskiy, German Ros, Felipe Codevilla, Antonio Lopez, and Vladlen
  Koltun.
\newblock Carla: An open urban driving simulator.
\newblock In {\em Conference on robot learning}, pages 1--16. PMLR, 2017.

\bibitem{drews2017aggressive}
Paul Drews, Grady Williams, Brian Goldfain, Evangelos~A Theodorou, and James~M
  Rehg.
\newblock Aggressive deep driving: Combining convolutional neural networks and
  model predictive control.
\newblock In {\em Conference on Robot Learning}, pages 133--142. PMLR, 2017.

\bibitem{fisac2018general}
Jaime~F Fisac, Anayo~K Akametalu, Melanie~N Zeilinger, Shahab Kaynama, Jeremy
  Gillula, and Claire~J Tomlin.
\newblock A general safety framework for learning-based control in uncertain
  robotic systems.
\newblock {\em IEEE Transactions on Automatic Control}, 64(7):2737--2752, 2018.

\bibitem{fisac2019bridging}
Jaime~F Fisac, Neil~F Lugovoy, Vicen{\c{c}} Rubies-Royo, Shromona Ghosh, and
  Claire~J Tomlin.
\newblock Bridging hamilton-jacobi safety analysis and reinforcement learning.
\newblock In {\em 2019 International Conference on Robotics and Automation
  (ICRA)}, pages 8550--8556. IEEE, 2019.

\bibitem{fuchs2020superhuman}
F. Florian, S. Yunlong, E. Kaufmann, D. Scaramuzza, and P. Duerr.
\newblock Super-human performance in gran turismo sport using deep
  reinforcement learning, 2020.

\bibitem{francis2021core}
Jonathan Francis, Nariaki Kitamura, Felix Labelle, Xiaopeng Lu, Ingrid Navarro,
  and Jean Oh.
\newblock Core challenges in embodied vision-language planning.
\newblock {\em arXiv preprint arXiv:2106.13948}, 2021.

\bibitem{garcia2015comprehensive}
Javier Garc{\i}a and Fernando Fern{\'a}ndez.
\newblock A comprehensive survey on safe reinforcement learning.
\newblock {\em Journal of Machine Learning Research}, 16(1):1437--1480, 2015.

\bibitem{optimized_dp}
George Giovanis, Michael Lu, and Mo Chen.
\newblock Optimizing dynamic programming-based algorithms.
\newblock \url{https://github.com/SFU-MARS/optimized_dp}, 2021.

\bibitem{grigorescu2020survey}
Sorin Grigorescu, Bogdan Trasnea, Tiberiu Cocias, and Gigel Macesanu.
\newblock A survey of deep learning techniques for autonomous driving.
\newblock {\em Journal of Field Robotics}, 37(3):362--386, 2020.

\bibitem{haarnoja2018soft}
Tuomas Haarnoja, Aurick Zhou, Pieter Abbeel, and Sergey Levine.
\newblock Soft actor-critic: Off-policy maximum entropy deep reinforcement
  learning with a stochastic actor.
\newblock In {\em International Conference on Machine Learning}, pages
  1861--1870. PMLR, 2018.

\bibitem{herman2021learn}
James Herman, Jonathan Francis, Siddha Ganju, Bingqing Chen, Anirudh Koul,
  Abhinav Gupta, Alexey Skabelkin, Ivan Zhukov, Max Kumskoy, and Eric Nyberg.
\newblock Learn-to-race: A multimodal control environment for autonomous
  racing.
\newblock In {\em Proceedings of the IEEE/CVF International Conference on
  Computer Vision}, pages 9793--9802, 2021.

\bibitem{hsu2021safety}
Kai-Chieh Hsu, Vicen{\c{c}} Rubies-Royo, Claire~J Tomlin, and Jaime~F Fisac.
\newblock Safety and liveness guarantees through reach-avoid reinforcement
  learning.
\newblock In {\em Robotics: Science and Systems}, 2021.

\bibitem{kabzan2019learning}
Juraj Kabzan, Lukas Hewing, Alexander Liniger, and Melanie~N Zeilinger.
\newblock Learning-based model predictive control for autonomous racing.
\newblock {\em IEEE Robotics and Automation Letters}, 4(4):3363--3370, 2019.

\bibitem{kingma2014adam}
Diederik~P Kingma and Jimmy Ba.
\newblock Adam: A method for stochastic optimization.
\newblock {\em arXiv preprint arXiv:1412.6980}, 2014.

\bibitem{kingma2013auto}
Diederik~P Kingma and Max Welling.
\newblock Auto-encoding variational bayes.
\newblock {\em arXiv preprint arXiv:1312.6114}, 2013.

\bibitem{kong2015kinematic}
Jason Kong, Mark Pfeiffer, Georg Schildbach, and Francesco Borrelli.
\newblock Kinematic and dynamic vehicle models for autonomous driving control
  design.
\newblock In {\em 2015 IEEE Intelligent Vehicles Symposium (IV)}, pages
  1094--1099. IEEE, 2015.

\bibitem{kousik2020bridging}
Shreyas Kousik, Sean Vaskov, Fan Bu, Matthew Johnson-Roberson, and Ram
  Vasudevan.
\newblock Bridging the gap between safety and real-time performance in
  receding-horizon trajectory design for mobile robots.
\newblock {\em The International Journal of Robotics Research},
  39(12):1419--1469, 2020.

\bibitem{kumar2020conservative}
Aviral Kumar, Aurick Zhou, George Tucker, and Sergey Levine.
\newblock Conservative q-learning for offline reinforcement learning.
\newblock {\em arXiv preprint arXiv:2006.04779}, 2020.

\bibitem{liniger2015optimization}
Alexander Liniger, Alexander Domahidi, and Manfred Morari.
\newblock Optimization-based autonomous racing of 1: 43 scale rc cars.
\newblock {\em Optimal Control Applications and Methods}, 36(5):628--647, 2015.

\bibitem{ohn2020learning}
Eshed Ohn-Bar, Aditya Prakash, Aseem Behl, Kashyap Chitta, and Andreas Geiger.
\newblock Learning situational driving.
\newblock In {\em Proceedings of the IEEE/CVF Conference on Computer Vision and
  Pattern Recognition}, pages 11296--11305, 2020.

\bibitem{prakash2021multi}
Aditya Prakash, Kashyap Chitta, and Andreas Geiger.
\newblock Multi-modal fusion transformer for end-to-end autonomous driving.
\newblock In {\em Proceedings of the IEEE/CVF Conference on Computer Vision and
  Pattern Recognition}, pages 7077--7087, 2021.

\bibitem{ray2019benchmarking}
Alex Ray, Joshua Achiam, and Dario Amodei.
\newblock Benchmarking safe exploration in deep reinforcement learning.
\newblock {\em arXiv preprint arXiv:1910.01708}, 7, 2019.

\bibitem{rosolia2017autonomous}
Ugo Rosolia, Ashwin Carvalho, and Francesco Borrelli.
\newblock Autonomous racing using learning model predictive control.
\newblock In {\em 2017 American Control Conference (ACC)}, pages 5115--5120.
  IEEE, 2017.

\bibitem{schulman2017proximal}
John Schulman, Filip Wolski, Prafulla Dhariwal, Alec Radford, and Oleg Klimov.
\newblock Proximal policy optimization algorithms.
\newblock {\em arXiv preprint arXiv:1707.06347}, 2017.

\bibitem{srinivasan2020learning}
Krishnan Srinivasan, Benjamin Eysenbach, Sehoon Ha, Jie Tan, and Chelsea Finn.
\newblock Learning to be safe: Deep {RL} with a safety critic.
\newblock {\em arXiv preprint arXiv:2010.14603}, 2020.

\bibitem{strobel2020accurate}
Kieran Strobel, Sibo Zhu, Raphael Chang, and Skanda Koppula.
\newblock Accurate, low-latency visual perception for autonomous racing:
  Challenges, mechanisms, and practical solutions.
\newblock In {\em 2020 IEEE/RSJ International Conference on Intelligent Robots
  and Systems (IROS)}, pages 1969--1975. IEEE, 2020.

\bibitem{thananjeyan2021recovery}
Brijen Thananjeyan, Ashwin Balakrishna, Suraj Nair, Michael Luo, Krishnan
  Srinivasan, Minho Hwang, Joseph~E Gonzalez, Julian Ibarz, Chelsea Finn, and
  Ken Goldberg.
\newblock Recovery {RL}: Safe reinforcement learning with learned recovery
  zones.
\newblock {\em IEEE Robotics and Automation Letters}, 6(3):4915--4922, 2021.

\bibitem{voloshin2019empirical}
Cameron Voloshin, Hoang~M Le, Nan Jiang, and Yisong Yue.
\newblock Empirical study of off-policy policy evaluation for reinforcement
  learning.
\newblock {\em arXiv preprint arXiv:1911.06854}, 2019.

\bibitem{weiss2020deepracing}
Trent Weiss and Madhur Behl.
\newblock Deepracing: a framework for autonomous racing.
\newblock In {\em 2020 Design, Automation \& Test in Europe Conference \&
  Exhibition (DATE)}, pages 1163--1168. IEEE, 2020.

\bibitem{yang2020projection}
Tsung-Yen Yang, Justinian Rosca, Karthik Narasimhan, and Peter~J Ramadge.
\newblock Projection-based constrained policy optimization.
\newblock {\em arXiv preprint arXiv:2010.03152}, 2020.

\bibitem{yurtsever2020survey}
Ekim Yurtsever, Jacob Lambert, Alexander Carballo, and Kazuya Takeda.
\newblock A survey of autonomous driving: Common practices and emerging
  technologies.
\newblock {\em IEEE access}, 8:58443--58469, 2020.

\bibitem{zhang2021learning}
Jimuyang Zhang and Eshed Ohn-Bar.
\newblock Learning by watching.
\newblock In {\em Proceedings of the IEEE/CVF Conference on Computer Vision and
  Pattern Recognition}, pages 12711--12721, 2021.

\bibitem{zhang2021end}
Zhejun Zhang, Alexander Liniger, Dengxin Dai, Fisher Yu, and Luc Van~Gool.
\newblock End-to-end urban driving by imitating a reinforcement learning coach.
\newblock In {\em Proceedings of the IEEE/CVF International Conference on
  Computer Vision}, pages 15222--15232, 2021.

\end{thebibliography}
}
\appendix
\counterwithin{figure}{section}
\counterwithin{table}{section}
\counterwithin{equation}{section}
\newpage
\section{Classical Control Benchmarks} 
The objective of this section is to compare the learning rule proposed by \cite{fisac2019bridging}, i.e.,  $Q(x, u) = (1-\gamma)l(x) + \gamma\min  \{l(x), \max_{u'\in \mathcal{U}}Q(x', u'\}$
with alternatives for learning safety value function. We evaluate it on two classical control benchmarks, \textit{Double Integrator} and \textit{Dubins' Car}, as described in Section \ref{sec:toy_model}, where the analytical solution to safe states and optimal safe actions are known. Thus, we implement the learning rule here as Eqn. \ref{eq:lr_known_safety}. This is slightly different from the general case of Eqn. \ref{eqn:safey_update}, where the optimal safety policy is unknown.
\begin{multline}\label{eq:lr_known_safety}
Q(x, u) = (1-\gamma)l(x) + \gamma\min  \{l(x), Q(x', u'\},\\
 \text{where}\; u' = \pi^*_S.
\end{multline}  

\subsection{Model Dynamics} \label{sec:toy_model}

\para{Double Integrator.} The double integrator models a particle moving along the x-axis at velocity $v$. The control input is the acceleration $a$. The goal in this case is keep the particle within a fixed boundary, in this case $x\in [-1, 1]$, subject to $a\in [-1, 1]$. 
\begin{equation}
\begin{cases}
\dot{x}=v\\
\dot{v}=a
\end{cases}
\end{equation}
By solving the Hamiltonian, i.e., $\pi_S^*(x) = \arg \max_{u \in \mathcal{U}} \langle f(x, u), \nabla V_S(x) \rangle$, we can get the optimal safe control as:
\begin{equation}\label{eq:opt_DI}
  a^*=
 \begin{cases}
     \overline{a} & \text{if}\;\; \partial V_S/\partial v \geq  0\\
     \underline{a} & \text{otherwise}   
 \end{cases} 
\end{equation}
\para{Dubins' Car.} The Dubins' car models a vehicle moving at constant speed, in this case $v=1$. Similar to the kinematic vehicle model, $x, y, \phi$ describes the position and heading of the vehicle, and control input is the turning rate $u\in[-1, 1]$. The goal is to reach a unit circle centred at the origin.   
\begin{equation}
\begin{cases}
\dot{x}=v\cos(\phi)\\
\dot{y}=v\sin(\phi)\\
\dot{\phi}=u
\end{cases}
\end{equation}
Note that Dubins' Car is a reach task, i.e. reaching a specified goal, instead of an avoid task, i.e. avoiding specified obstacles. The reach task can be simply implemented by setting  $\pi_S^*(x) = \arg \min_{u \in \mathcal{U}} \langle f(x, u), \nabla V_S(x) \rangle$ \citep{bansal2017hamilton}. In other words, the optimal safe action for a given state is the one that minimises the distance to the goal. The corresponding optimal safe control is 
\begin{equation}
  u^*=
 \begin{cases}
     \underline{u} & \text{if}\;\; \partial V_S/\partial \theta \geq  0\\
     \overline{u} & \text{otherwise}   
 \end{cases} 
\end{equation}

The ground truth safety value function for these two benchmarks are shown in Figure \ref{fig:DoubleIntegrator} and \ref{fig:DubinsCar}.

\para{Implementation \& Evaluation.} We use a neural network with hidden layers of size [16, 16] for the double integrator and [64, 64, 32] for  Dubins' car. We use ADAM \cite{kingma2014adam} as the optimiser with a learning rate of 0.001, batch size of 64. We update the safety value function over 25K steps for Double Integrator and 50K steps for Dubin's Car, and report classification accuracy every 1000 steps averaged over 5 random seeds. While the safety value is defined over continuous state space, we evaluate the performance over a discrete mesh on the state space. By definition, the safety value at a given state $x$ is $Q(x, u^*)$, where $u^*=\pi^*_S(x)$.  

\begin{figure}[h!]
    \centering
    \includegraphics[width=\linewidth]{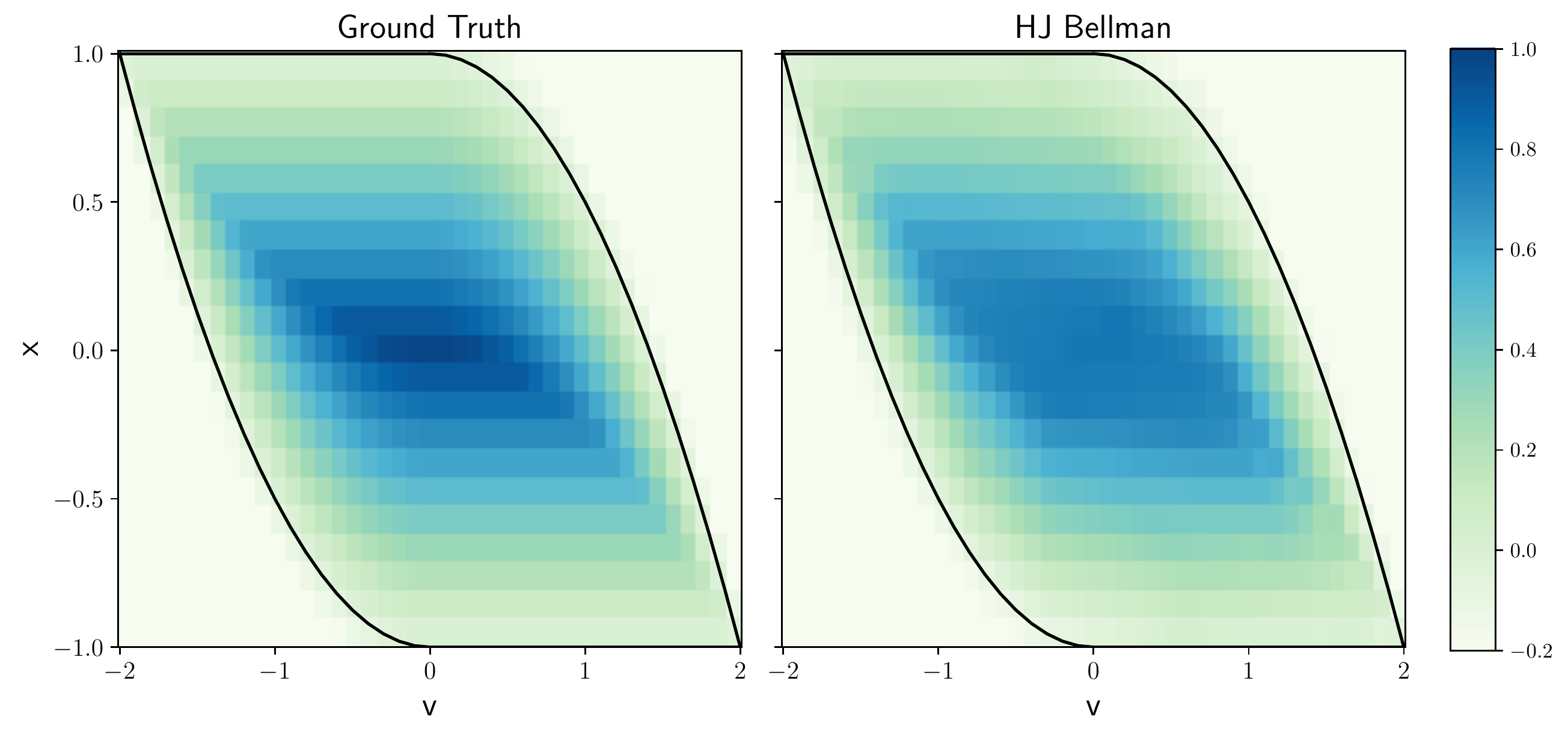}
    \caption{A comparison between the ground truth safety value and that learned via HJ Bellman update for double integrator; The black line delineates $V_S(x)=0$.}
    \label{fig:V_est_DI}
\end{figure}

\begin{figure}[h!]
     \centering
     \begin{subfigure}[b]{0.22\textwidth}
         \centering
         \includegraphics[clip, trim=  5cm 2cm 5cm 2cm, width=\linewidth]{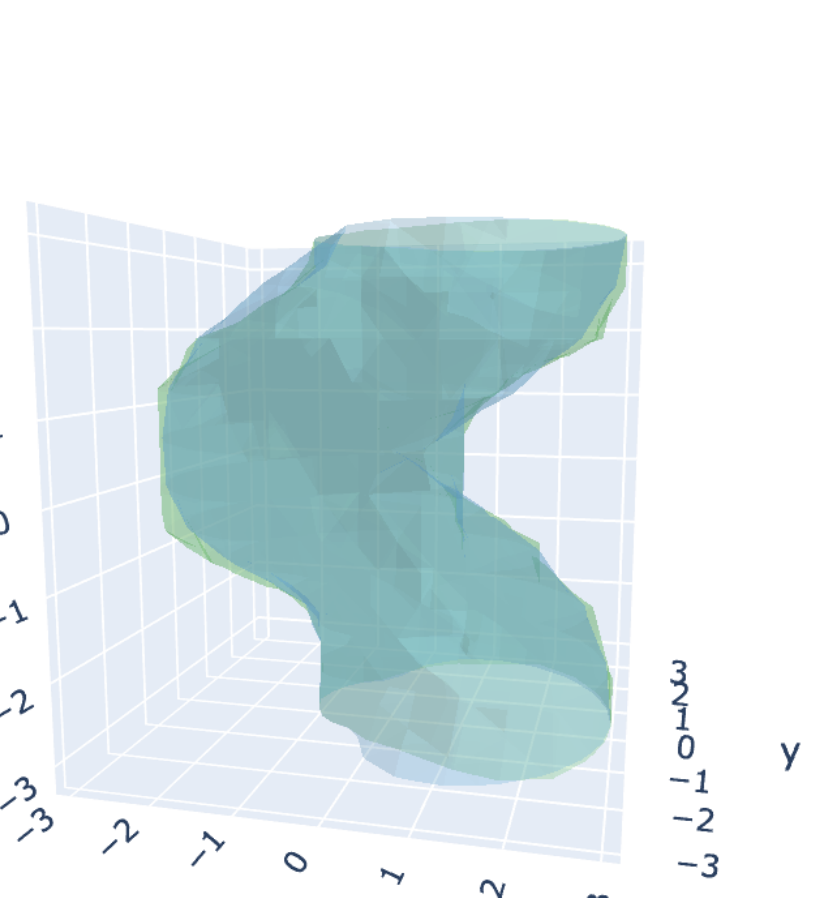}
         \caption{}
         \label{fig:DC}
     \end{subfigure}
     \quad
     \begin{subfigure}[b]{0.22\textwidth}
         \centering
         \includegraphics[clip, trim= 5cm 2cm 5cm 2cm, width=\linewidth]{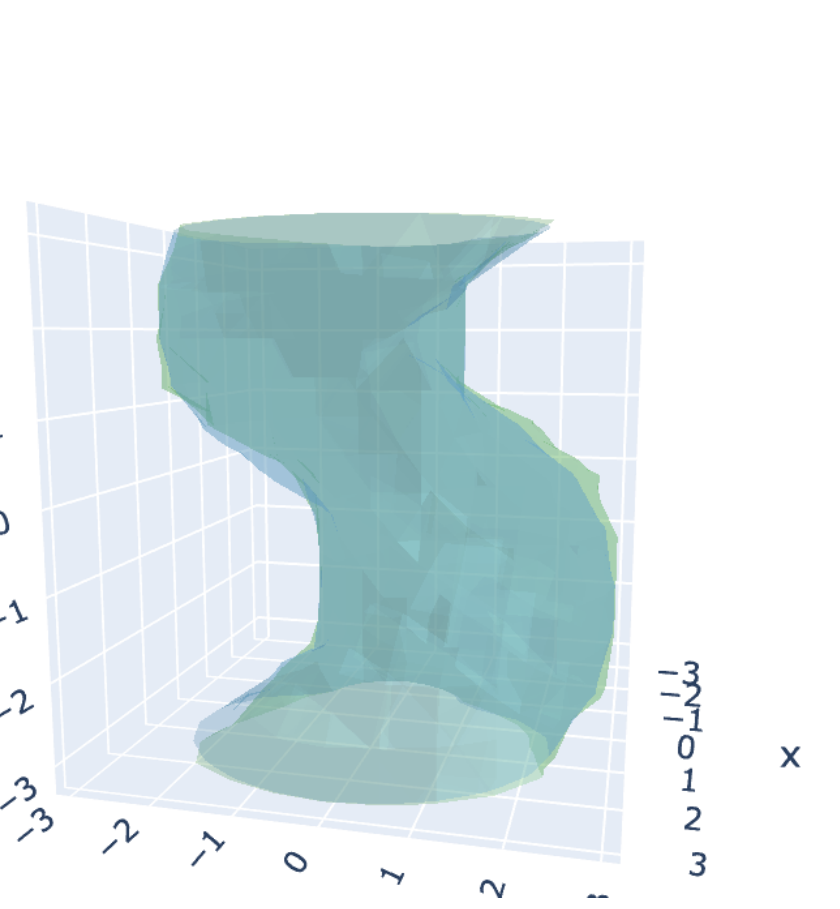}
         \caption{}
         \label{fig:Dubins}
     \end{subfigure}
        \caption{A comparison between isosurface of the ground truth safety value (blue) and that learned via HJ Bellman update (green) for Dubins' car }
        \label{fig:V_est_Dubins}
\end{figure}

\para{Qualitative Results.} Qualitative comparison between the ground truth value and that learned via HJ Bellman update is shown in Figure \ref{fig:V_est_DI} and  \ref{fig:V_est_Dubins}. As we can see, the neural approximation largely recovers the ground truth value, except for minute difference.


\begin{figure}[h!]
    \centering
    \includegraphics[width=0.75\linewidth]{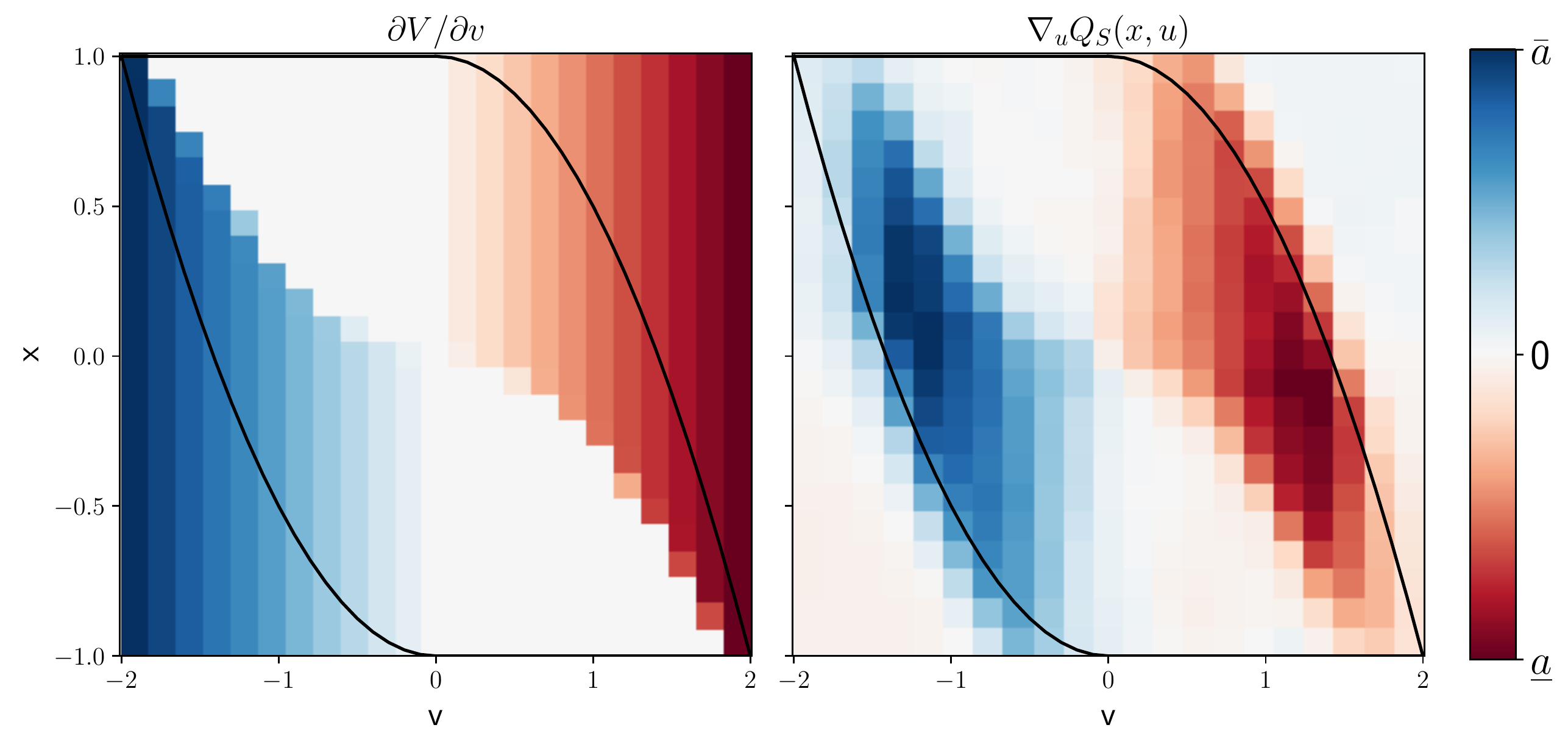}
    \caption{The gradients through the safety critic, i.e., $\nabla_u Q_S(x, u)$, consistently point towards the correct optimal safe action, as indicated by $\partial{V}/\partial{v}$ (Eqn. \ref{eq:opt_DI}), within the safe set (the area delineated by the black line) for double integrator.}
    \label{fig:nabla_Q}
\end{figure}

While we do not need to learn the safety actor in this case, we further demonstrate that $\nabla_u Q_S(x, u)$ can indeed by used to update the safety actor. In Figure \ref{fig:nabla_Q}, we compare the ground truth $\partial{V}/\partial{v}$, with which one can determine the optimal safe action with Eqn. \ref{eq:opt_DI}, and the gradient through the safety critic, i.e., $\nabla_u Q_S(x, u)$. We can see that $\nabla_u Q_S(x, u)$, consistently point towards the correct optimal safe action within the safe set, i.e., the area delineated by the black line. The safety value outside the safe set is not learned, as the region is outside the support of data when episodes terminate upon failures. 

\subsection{Learning Rule Comparison for Safety Critic }\label{sec:toy_comp}
Firstly, we describes the approaches pertaining to learning the safety critic in \cite{srinivasan2020learning, bharadhwaj2020conservative}. 
In both Safety Q-functions for RL (SQRL) \citep{srinivasan2020learning} and Conservative Safety Critic (CSC) \citep{bharadhwaj2020conservative}, the safety critic is defined as the expected cumulative cost, i.e. $Q^\pi_C(x, u) \coloneqq \mathbb{E}_{x_k, u_k \sim \pi} \left[\sum_{k=0}^\infty \gamma^k C(x_k)|x_0=x, u_0=u\right]$, where $C(x_k)=1$ if a failure occurs at $x_k$ and 0 otherwise. Both papers endowed the safety critic, $Q^\pi_C(x, u)$ with a probabilistic interpretation, i.e. the expected probability of failure.

\para{SQRL.} The safety critic is trained by propagating the failure signal using the standard Bellman backup, as in Eqn. \ref{eq:SQRL}, where $\mathcal{D}$ denotes the replay memory, $\gamma_S$ is a time-discount parameter, and $\bar{Q}^\pi_C$ is the delayed target network. This approach for learning the safety critic is also adopted in \cite{thananjeyan2021recovery}.
\begin{equation}\label{eq:SQRL}
    Q_C^\pi \leftarrow {E}_{x, u, x'\sim \mathcal{D}}\left[C(x) + \gamma_S(1-C(x)) \mathbb{E}_{u'\sim \pi} \bar{Q}^\pi_C(x', u')\right]
\end{equation}
\para{CSC.} On top of using Bellman backup to propagate the failure signals, CSC uses conservative Q-learning (CQL) \citep{kumar2020conservative} to correct for the distribution mismatch between the behaviour policy and the evaluation policy, and overestimate $Q_C$ to err on the side of caution. The resulting objective is given in Eqn. \ref{eq:CQL}, where $\mathcal{B}^\pi Q(x, u)=C(x)+\gamma\mathbb{E}_{x'\sim P(x'|x, u), u'\sim \pi(x')} \bar{Q}(x', u')$ is the Bellman operator and $\alpha$ is a hyperparamter that controls the extent of conservativeness. If $\alpha =0$, the objective is the same as that of SQRL.
\begin{equation}\label{eq:CQL}
\begin{split}
    \mathcal{L}&= \frac{1}{2} \mathbb{E}_{s,a\sim \mathcal{D}}\left[Q_C(x, u) -\mathcal{B}^\pi \bar{Q}_C(x, u))\right] \\&-\alpha \left[ \mathbb{E}_{x\sim\mathcal{D}, u\sim \pi(x)}Q_C(x, u)-\mathbb{E}_{x, u\sim \mathcal{D}}Q_C(x, u)\right]
    \end{split}
\end{equation}
Note that CSC reversed the sign in front of $\alpha$ compared to the original implementation in CQL so as to over-estimate $Q_C$. This learning objective does not guarantee point-wise conservativeness, but conservativeness in expectation, i.e. $\mathbb{E}_\pi  \hat{Q}_c^\pi(x, u)\geq \mathbb{E}_\pi  {Q}_c^\pi(x, u)$. $\tau$ is the rate for polyak averaging of the target network, i.e.  $Q' \xleftarrow{} \tau Q + (1-\tau)Q'$.

\para{Implementation Details.} In \cite{srinivasan2020learning}, the authors used a learning rate of $3\times10^{-4}$, $\gamma_S=0.7$. Using the same learning rate, we did grid search over $\gamma_S=[0.7, 0.9]$ and $\tau =[0.1, 0.01]$. We observed that $\gamma_S=0.9$ had better performance, and thus selected $\gamma_S = 0.9$ and $\tau=0.1$. 

In \cite{bharadhwaj2020conservative}, the authors used a learning rate of $2\times10^{-4}$, $\gamma_S=0.99$, and selected $\alpha = 0.5$ from from 0.05, 0.5, and 5. Using the same learning rate and $\gamma_S$, we did grid search over $\alpha=[0.01, 0.05, 0.5, 5]$ and $\tau =[0.1, 0.01]$. We selected $\alpha = 0.01$ and $\tau=0.1$. 

\begin{figure}[h!]
    \centering
    \includegraphics[width=\linewidth]{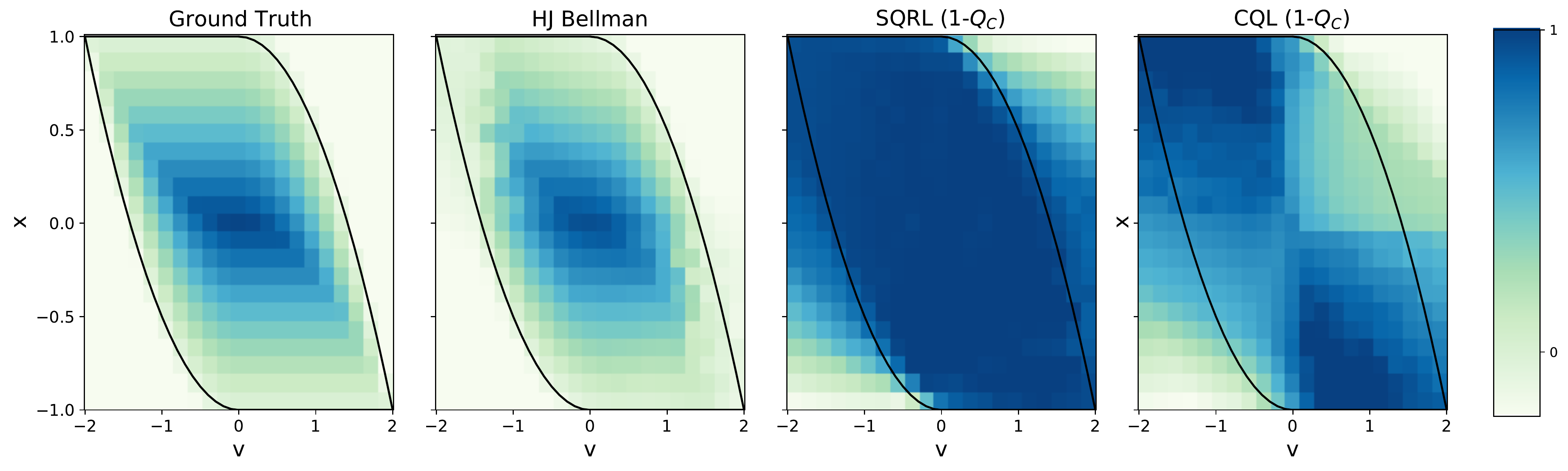}
    \caption{Comparison between the group truth safety value and the safety critics from different learning rules for double integrator}
    \label{fig:V_DI_comp}
\end{figure}

\para{Results.} The main results are summarized in Figure \ref{fig:toy_accuracy}. In Figure \ref{fig:V_DI_comp}, we show a qualitative comparison between the ground truth safety value and that learned via different learning rules. In interpreting the results, note that both the environment and optimal safety policy are deterministic. Thus, $Q^{\pi_S^*}_C(x, \pi_S^*(x))$ should be 0 if $x$ is a safe state, following the definition. Due to the difference in definition, i.e., $Q_S(x)\geq 0$ is safe for HJ safety value and $Q_C(x)\leq 0$ is safe in SQRL and CQL, we plot $1-Q_C$ such that in Figure \ref{fig:V_DI_comp} the larger value consistently indicates safety and the cut-off for safe vs. unsafe is 0.

SQRL largely captures the correct safe states, though the classification performance is  highly dependent on picking an appropriate threshold. CQL does underestimate the level of safety (and overestimate $Q_C$) as intended, and the pattern of underestimation appear to corresponds to $\nabla_x Q_S$  (refer to Figure \ref{fig:nabla_Q}). 

\newpage
\section{\ourmethod~Algorithm}\label{sec:algorithm}
\ourmethod~relies on a dual policy structure, the rationale of which is explained in Section \ref{sec:approach}.  This pairing of a safety policy and a performance policy is important, as we are able to decompose the problem of learning under safety constraints into optimizing for performance and updating the safety value function, separately. 

We optimize the performance policy using SAC, but it may be switched for any other comparable RL algorithms. The safety policy is used least-restrictively, that is only intervene when the RL agent is about to enter into an unsafe state and thus allowing the performance policy maximum freedom in exploring safely. Instead of using the optimal safe policy from solving Hamiltonian, the safe policy is updated via gradients through the safety critic, same as other actor-critic algorithms.  


\begin{algorithm}[h!] 
\SetAlgoLined
\textbf{Initialize: } performance critic $Q_{\phi}$ and actor $\pi_\theta$; \\
\textbf{Initialize: } safety critic $Q_{\phi_S}$, and actor $\pi_{\theta_S}$; target networks  $\phi'_S\leftarrow \phi_S$;\\ 
\textbf{Initialize: } replay buffer $\mathcal{D}$;\\
 \For{i = 0, $\dots$, \# Episodes}{
 $x $ = env.reset()\\
 \While{not terminal}{
 {u $\sim \pi_{\theta}(x)$;\\
 \emph{// The safe actor intervenes when the current state-action is deemed unsafe by the safety critic.}\\
  \If{$Q_{\phi_S}(x, u)< \epsilon$}{
  $u \sim \pi_{\theta_S}(x)$
   }
 $x', r $ = env.step($u$)\\
 $\mathcal{D}$.store($x, a, x', r$)\\
 $x=x'$\\
 {Update performance critic $Q_{\phi}$ and actor $\pi_\theta$ with preferred RL algorithm;\\
  \emph{// Update the safety critic:}\\
 Sample N transitions $(x, u, x')$ from $\mathcal{D}$;\\
 \emph{// Calculate the target value with the discounted Bellman safety update }
 $$y = (1-\gamma)l(x)+ \gamma \min\{l(x), Q_{\phi'_S}(x', u')\},$$ $$\text{where}\;u'\sim \pi_{\theta_S}$$
 $$\mathcal{L}_{\phi_S} = N^{-1}\sum (Q_{\phi_S}(x, u)-y)^2$$
 $$\phi_S \leftarrow\phi_S- \alpha\nabla_{\phi_S} \mathcal{L}_{\phi_S}$$
  \emph{// Update the safety actor with deterministic policy gradient:}
   $$\theta_S \leftarrow\theta_S + \alpha N^{-1}\sum \nabla_u Q(x, u) \nabla_{\theta_S} \pi_{\theta_S}(x)$$
   \emph{// Update the target networks:}
   $$\phi'_S\leftarrow \tau\phi_S+(1-\tau)\phi'_S $$ 
 }
}
 }}
 \caption{\texttt{\ourmethod}: Safe Autonomous Racing via Approximate Reachability on Ego-vision}
 \label{suppl:alg:spar}
\end{algorithm}


\section{Details: Safety Gym Experiment}
\label{app:sec:sgym_details}

Following the default CarGoal1-v0 and PointGoal1-v0 benchmarks in Safety Gym, all agents were given LiDARs observation with respect to hazard, goal, and vase, with avoiding hazards as the safety constraints. Both environments were initialised with a total of 8 hazards and 1 vase. Agent's are endowed with accelerometer, velocimeter, gyro, and magnetometer sensors; their LiDAR configurations included 16 bins, with max distance of 3.

The baselines we considered, i.e., CPO, PPO and PPO-Lagrangian follows the default implementation that comes with Safety Gym. PPO-\ourmethod~wraps the proposed safety actor critic around the PPO base agent. Despite PPO being an on-policy algorithm, the \ourmethod~safety critic was implemented with off-policy updates, using prioritised memory replay based on the TD-error of predicting safety value. Since $l(x)$ is small in this environment, we scaled cost by a factor of 100. For the safety actor-critic, We used $\gamma_S$ annealing from 0.85 to 1 following \cite{fisac2019bridging}, $\tau =  0.005$, critic learning rate of 0.001, actor learning rate of 0.0003, and $\alpha =0.2$ (regularisation on policy entropy). We used a safety margin $\epsilon = 0.25$, mainly to account for the dimension of the hazards (radius = 0.2). 

For each model, on each Safety Gym benchmark, results were reported as the average across 5 instances. All experiments in Safety Gym were run on an Intel(R) Core(TM) i9-9920X CPU @ 3.50GHz -- with 1 CPU,  12 physical cores per CPU, and a total of 24 logical CPU units.


\section{Static Safety Actor-Critic Derived from Kinematic Bike Model}\label{sec:nominal_ac}

 \begin{figure*}[h!]
 \vspace{-0.0cm}
     \centering
   \begin{subfigure}[b]{0.25\textwidth}
         \centering
 \includegraphics[width=0.8\textwidth]{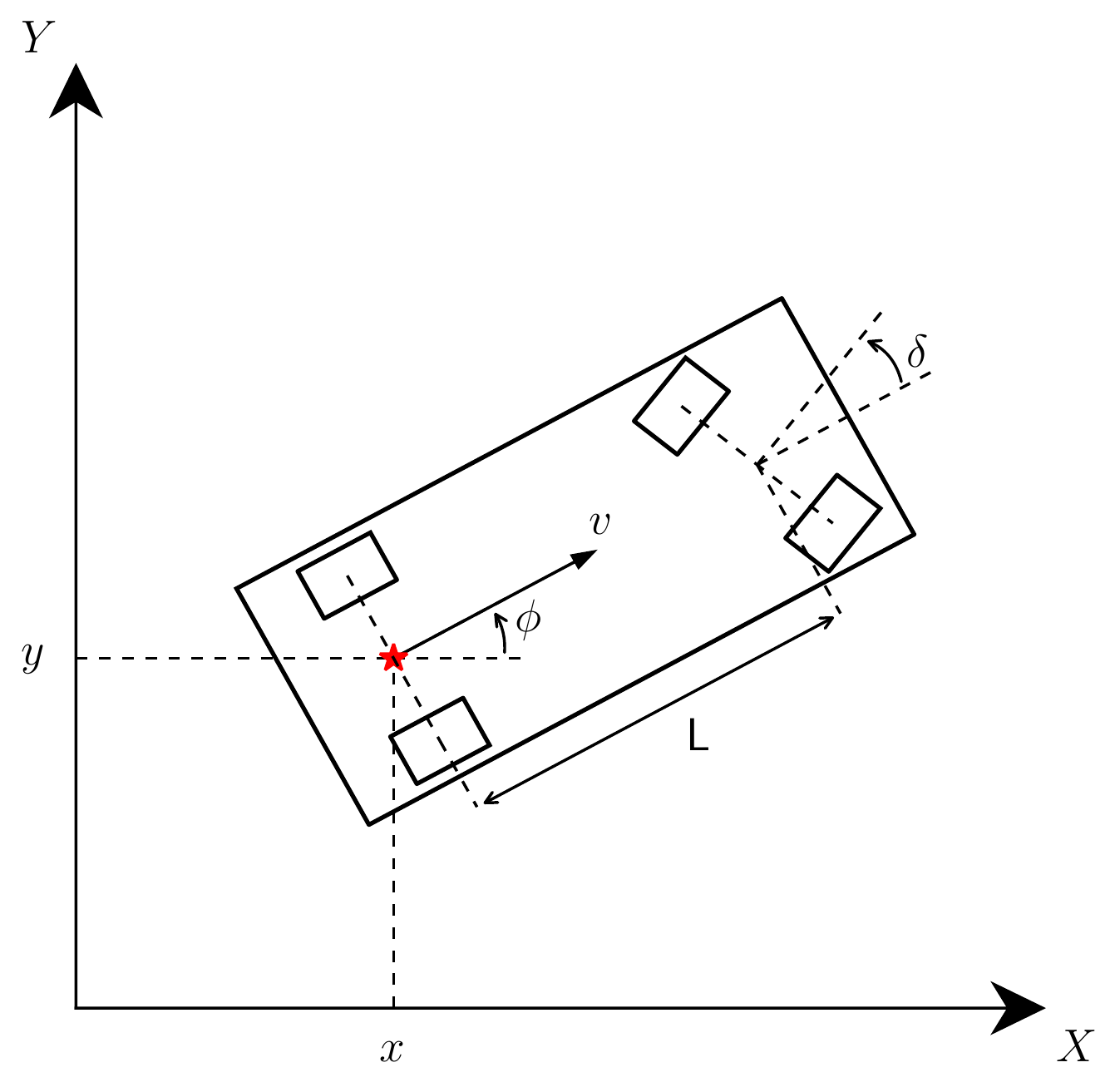}
 \caption{nominal model}
\label{fig:bike_model}
\end{subfigure}
\begin{subfigure}[b]{0.69\textwidth}
         \centering
 \includegraphics[width=\textwidth]{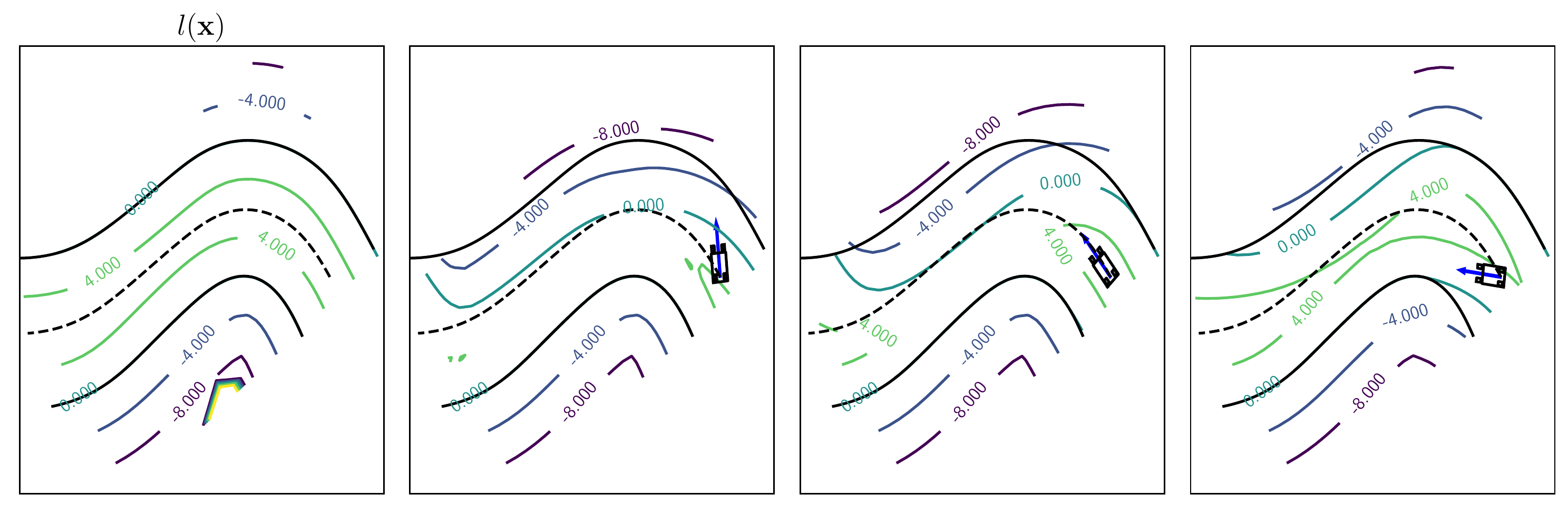}
 \caption{$V_s(x, y, v, \phi)$ computed via the nominal model, where v=12m/s}
\label{fig:HJ_contour}
\end{subfigure}
     \caption{(a) We compute the safety value function, via a kinematic vehicle model. (b) We illustrate different views of the 4D state space, given fixed velocity and three different yaw angles, indicated by the blue arrows.} 
     \label{fig:HJ_warmstart}
 \end{figure*}
 
To demonstrate the benefit of utilizing domain knowledge in the form of a nominal model and to compare with the learnable safety actor-critic in \texttt{\ourmethod}, we use the kinematic vehicle model \cite{kong2015kinematic} (see Figure \ref{fig:bike_model}), which is a significant simplification of a realistic race car model \cite{kabzan2019learning}, to compute the safety value and corresponding `optimal'\footnote{only with respect to the nominal model} safety controller.  The dynamics and `optimal' safety control is given in Eqn. \ref{eq:bike_model} and \ref{eq:safety_control}, where the state is $\mathbf{x}=[x, y, v, \phi]$, and the action is $\mathbf{u}=[a, \delta]$.  $x, y, v, \phi$ are the vehicle's location, speed, and yaw angle. $a$ is the acceleration, and $\delta$ is the steering angle. The actions are bounded, i.e., $a\in [\underline{a}, \overline{a}]$ and $\delta \in [\underline{\delta}, \overline{\delta}]$. $L=3$m is the car length.

\begin{equation}\label{eq:bike_model}
f(\mathbf{x}, \mathbf{u}) = \begin{cases}
\dot{x}=v\cos(\phi)\\
\dot{y}=v\sin(\phi)\\
\dot{v}=a\\
\dot{\phi}=v\tan{\delta}/L
\end{cases}
\end{equation}

\begin{equation}\label{eq:safety_control}
\begin{split}
  a^*&=
 \begin{cases}
     \underline{a} & \text{if}\;\; {\partial V_S}/{\partial v}\leq  0\\
     \overline{a} & \text{else}   
 \end{cases}, \\
 \delta^*&=
 \begin{cases}
     \overline{\delta} & \text{if}\;\; {\partial V_S}/{\partial \phi} \geq  0\\
     \underline{\delta} &  \text{else}
 \end{cases}
 \end{split}
\end{equation}

Intuitively, the `optimal' safety policy  brakes and steers towards the center of the track as much as possible. We also derive the `optimal' safety policy here. The optimal safety control is derived by solving the Hamiltonian as given in Eqn. \ref{eq:hamiltonian}. By definition, $\nabla V_S(\mathbf{x}) = [{\partial V_S}/{\partial x }, {\partial V_S}/{\partial y}, {\partial V_S}/{\partial v}, {\partial V_S}/{\partial \phi}].$
\begin{subequations}\label{eq:derivation}
\begin{align}
    \pi_S^*(\mathbf{x}) &= \arg \max_{\mathbf{u} \in \mathcal{U}}  \langle f(\mathbf{x}, \mathbf{u}), \nabla V_S(\mathbf{x}) \rangle \label{eq:hamiltonian}\\
    &= \arg \max_{[a, \delta] \in \mathcal{U}} [v\cos(\phi)\frac{\partial V_S}{\partial x } + v\sin(\phi) \frac{\partial V_S}{\partial y } \nonumber\\
    &+ a \frac{\partial V_S}{\partial v} + v\tan{\delta}/L \frac{\partial V_S}{\partial \phi}]\\
    &= \arg \max_{[a, \delta] \in \mathcal{U}} [a \frac{\partial V_S}{\partial v}+ v\tan{\delta}/L \frac{\partial V_S}{\partial \phi}]\label{eq:derv_end}
\end{align}
\end{subequations}
From Eqn \ref{eq:derv_end}, it is clear that the actions given by Eqn. \ref{eq:safety_control} maximize the Hamiltonian.

We parametrized the racetrack as a cubic spline and computed $l(\mathbf{x})$ by projection onto the spline. 
Setting $V_S(\mathbf{x}, 0) = l(\mathbf{x})$, we calculated the backward reachable tube using the code from \cite{optimized_dp}. For efficient computation, we divided the racetrack into overlapping segments and computed the safety value segment-wise. 
Fig. \ref{fig:HJ_contour} illustrates resulting safety value function at slices of state space, as the agent enters into a sharp turn. It is clear that the safety value at each location can be quite different from the initialization, $l(\mathbf{x})$.



\section{Details: Learn-to-Race Experiment}\label{sec:implementation_details}
\para{Vision Encoder.} We condition the optimization of the performance policy as well as the safety value updates on pretrained embedding of vehicle's visual scene context. The perception module maps ego-images from the on-board RGB camera to feature embedding of reduced dimension. To learn this mapping, we use a standard variational autoencoding (VAE) \cite{kingma2013auto} paradigm, with a convolutional encoder. 

\begin{figure}[h!]
    \centering
    \includegraphics[width=\linewidth]{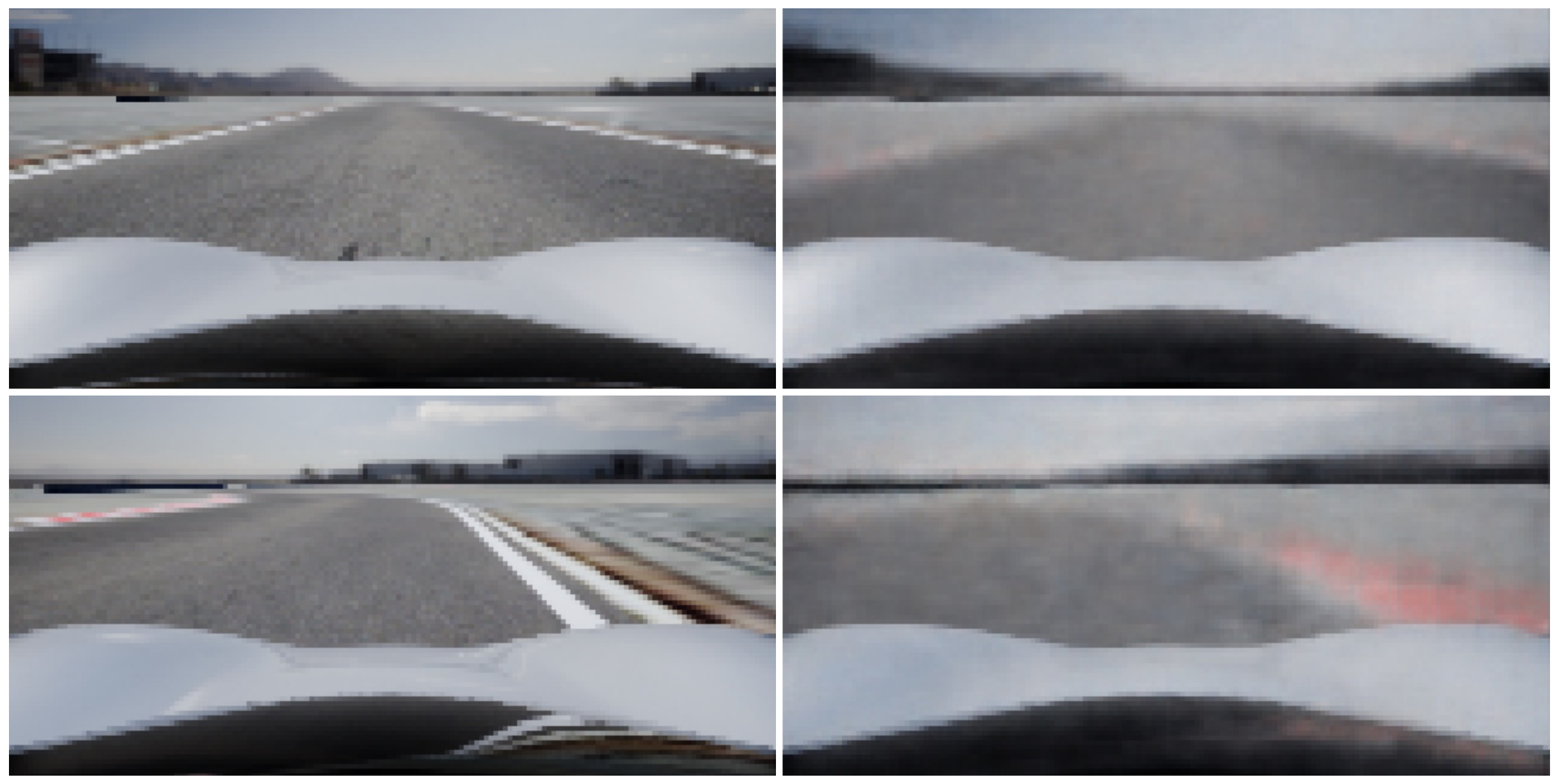}
    \caption{VAE image reconstruction, with real images in the left column and reconstructed images in the right column.}
    \label{fig:img_reconstruction}
\end{figure}

\begin{figure}[h!]
    \centering
    \includegraphics[width=\linewidth]{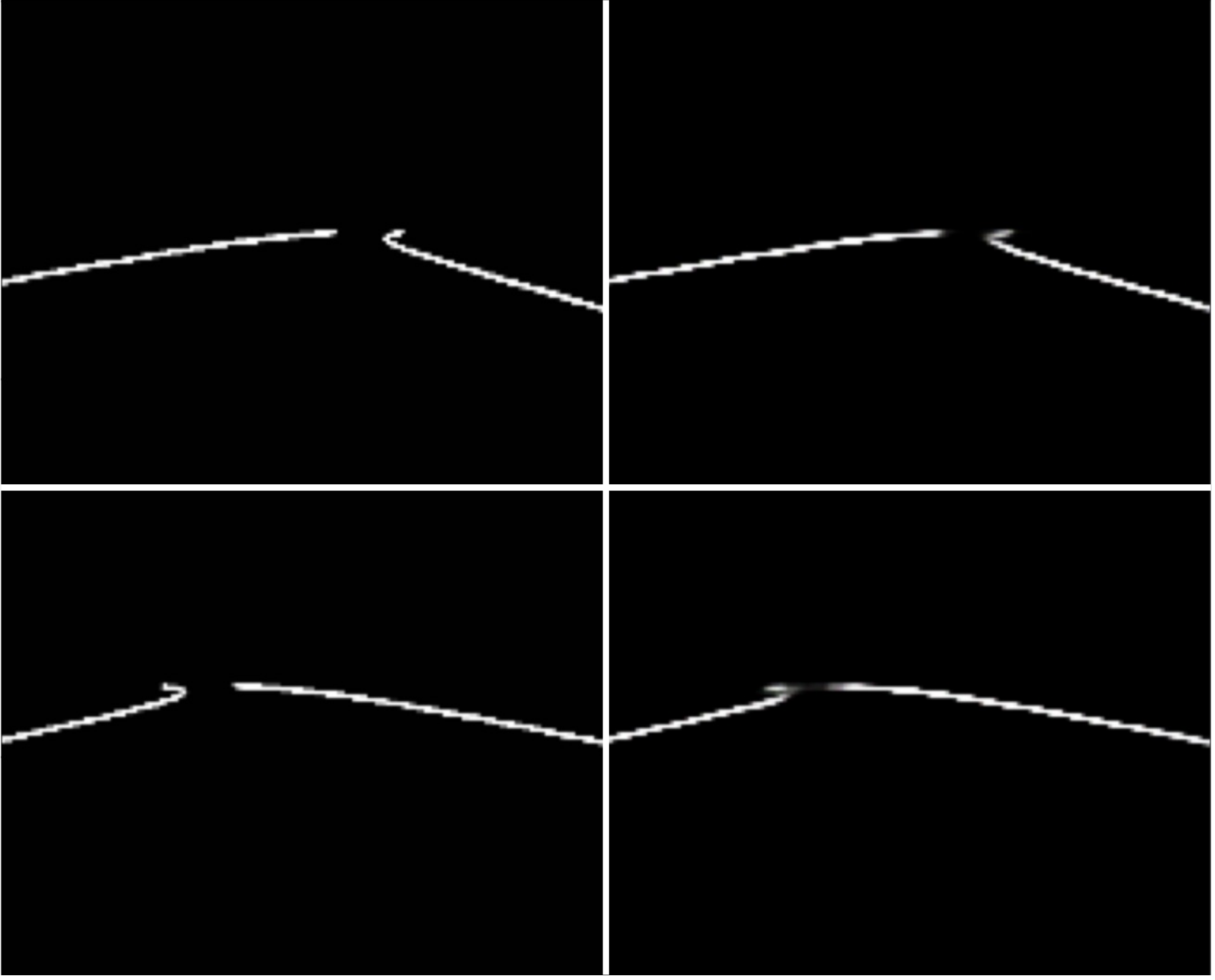}
    \caption{VAE reconstruction of projected road boundary images, with real images in the left column and reconstructed images in the right column.}
    \label{fig:road_boundaries}
\end{figure}

We use an image reconstruction objective with binary cross-entropy loss, Adam optimizer \cite{kingma2014adam}, and a latent vector dimension of 32. We train the VAE encoder to reconstruct ego-images, sampled from the vehicle's front camera during random agent execution; examples are provided in Figure \ref{fig:img_reconstruction}. We further refine the encoder by training the VAE module to reconstruct projected road boundaries, illustrated in Figure \ref{fig:road_boundaries}, with inputs in the left column and the reconstructed outputs in the right column.

\begin{figure}[h!]
    \centering
    \includegraphics[width=\linewidth]{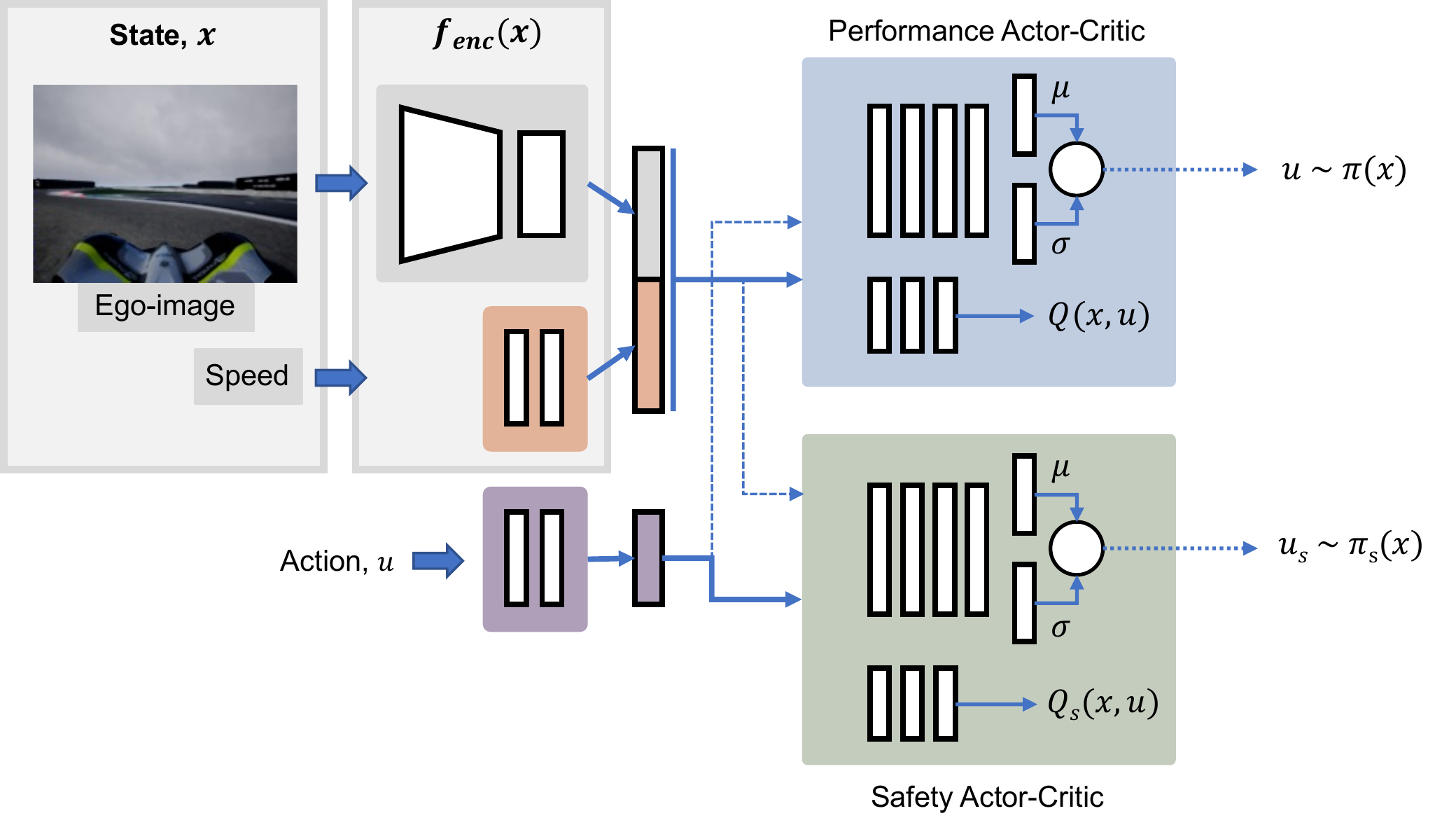}
    \caption{SAGE neural architecture overview.}
    \label{fig:enc_arch}
\end{figure}

\para{Neural Architecture.} As illustrated in Figure \ref{fig:enc_arch}, the vision encoder takes  an image as input to produce a latent vector, which is concatenated with speed and action embedding and passed to the performance and safety actor-critics. The specific implementation of layers are summarized in Table \ref{suppl:tab:hyperparameters}.

Specifically, we use a squashed Gaussian policy (Eqn. \ref{eq:squashed_gaussian}) for both performance and safety actors, following \cite{haarnoja2018soft}.
\begin{equation}\label{eq:squashed_gaussian}
    u = \tanh(\mu(x)+\sigma(x)\odot \xi),\quad \xi\sim \mathcal{N}(0, \mathbf{I})
\end{equation}
\para{Agent training details.} During training, the agent is spawned at random locations along the race track and uses a stochastic policy. During evaluation, the agent is spawned at a fixed location and uses a deterministic policy. The episode terminates when the agent successfully finishes a lap, leaves the drivable area, collides with obstacles, or does not progress for a number of steps. For each agent, we report averaged results across 5 random seeds, evaluated every 5000 steps over an episode (one lap). In total, we train each agent over 250,000 steps, and evaluate it over 50 episodes. 

During its interaction with the environment, the agent receives a $192 \times 144$ ego-camera view and its speed at each time-step. The agent encodes the RGB image frame and its speed to a 40-dimensional feature representation, subsequently used as input to both actor-critic networks. We initialise the replay buffer with 2000 random transitions, following \cite{SpinningUp2018}. After 2000 steps, we perform a policy update at each time step. For the SafeSAC agent, we only save state-action transitions from the performance actor to the replay buffer. For the \ourmethod~agent, we save all state-action transitions.

\begin{table*}[h!]
\caption{Network Architecture}
\label{suppl:tab:hyperparameters}
\begin{center}
\scriptsize
\resizebox{0.9\textwidth}{!}{
\begin{tabular}{l c l c l c l}
\toprule
Operation & & Input (dim.) & & Output (dim.) & & Parameters\\ 
\midrule
\multicolumn{7}{c}{\textsc{Visual Encoder}} \\
\hline
\textit{Conv2d} & & $(N, \text{chan}, 42, 144)$, chan : 3$\rightarrow$32 & & \texttt{conv1} & & k$\coloneqq$(4,4), s$\coloneqq$2, p$\coloneqq$1, activation$\coloneqq$ReLU \\
\textit{Conv2d} & & \texttt{conv1}, chan : 32$\rightarrow$64 & & \texttt{conv2} & & k$\coloneqq$(4,4), s$\coloneqq$2, p$\coloneqq$1, activation$\coloneqq$ReLU \\
\textit{Conv2d} & & \texttt{conv2}, chan : 64$\rightarrow$128 & & \texttt{conv3} & & k$\coloneqq$(4,4), s$\coloneqq$2, p$\coloneqq$1, activation$\coloneqq$ReLU \\
\textit{Conv2d} & & \texttt{conv3}, chan : 128$\rightarrow$256 & & \texttt{conv4} & & k$\coloneqq$(4,4), s$\coloneqq$2, p$\coloneqq$1, activation$\coloneqq$ReLU \\
\textit{Flatten} & & --- & & --- & & --- \\
\midrule
\multicolumn{7}{c}{\textsc{Visual Encoder Bottleneck Representation}} \\
\textit{Linear} (mu) & & $N \times \text{h\_dim}$ & & $N \times 32$ & & --- \\
\textit{Linear} (sigma) & & $N \times \text{h\_dim}$ & & $N \times 32$ & & --- \\
\midrule
\multicolumn{7}{c}{\textsc{Visual Decoder} (only for pre-training Visual Encoder)} \\
\hline
\textit{Unflatten} & & --- & & --- & & --- \\
\textit{ConvTranspose2d} & & \texttt{encoder.conv4}: encoder.conv4.chan: 256 $\rightarrow$128 & & \texttt{convtranspose1} & & k$\coloneqq$(4,4), s$\coloneqq$2, p$\coloneqq$1, activation$\coloneqq$ReLU \\
\textit{ConvTranspose2d} & & \texttt{convtranspose1}, chan : 128 $\rightarrow$64 & & \texttt{convtranspose2} & & k$\coloneqq$(4,4), s$\coloneqq$2, p$\coloneqq$1, activation$\coloneqq$ReLU \\
\textit{ConvTranspose2d} & & \texttt{convtranspose2}, chan : 64 $\rightarrow$32 & & \texttt{convtranspose3} & & k$\coloneqq$(4,4), s$\coloneqq$2, p$\coloneqq$1, activation$\coloneqq$ReLU \\
\textit{ConvTranspose2d} & & \texttt{convtranspose3}, chan : 32 $\rightarrow$3 & & \texttt{convtranspose4} & & k$\coloneqq$(4,4), s$\coloneqq$2, p$\coloneqq$1, activation$\coloneqq$Sigmoid \\
\midrule
\multicolumn{7}{c}{\textsc{Safety Actor-Critic}} \\
\hline
\texttt{actor\_network} & & --- & & --- & & --- \\
\texttt{q\_function1} & & --- & & --- & & --- \\
\texttt{q\_function2} & & --- & & --- & & --- \\
\midrule
\multicolumn{7}{c}{\textsc{Performance Actor-Critic}} \\
\hline
\texttt{actor\_network} & & --- & & --- & & --- \\
\texttt{q\_function1} & & --- & & --- & & --- \\
\texttt{q\_function2} & & --- & & --- & & --- \\
\midrule
\multicolumn{7}{c}{\textsc{Actor Network (Policy): SquashedGaussianMLPActor}} \\
\hline
\textit{Linear} & & $N \times 32$ & & $N \times 64$ & & activation$\coloneqq$ReLU \\
\textit{Linear} & & $N \times 64$ & & $N \times 64$ & & activation$\coloneqq$ReLU \\
\textit{Linear} & & $N \times 64$ & & $N \times 32$ & & activation$\coloneqq$ReLU \\
\textit{Linear} (projection: mu\_layer) & & $N \times 32$ & & $N \times 3$ & & --- \\
\textit{Linear} (projection: log\_std\_layer) & & $N \times 32$ & & $N \times 3$ & & --- \\
\midrule
\multicolumn{7}{c}{\textsc{Q function}} \\
\hline
\texttt{speed\_encoder} & & --- & & --- & & --- \\
\texttt{regressor} & & --- & & --- & & --- \\
\midrule
\multicolumn{7}{c}{\textsc{Speed Encoder}} \\
\hline
\textit{Linear} & & $N \times 1$ & & $N \times 8$ & & activation$\coloneqq$ReLU \\
\textit{Linear} & & $N \times 8$ & & $N \times 8$ & & activation$\coloneqq$Identity \\
\midrule
\multicolumn{7}{c}{\textsc{Regressor}} \\
\hline
\textit{Linear} & & $N \times 42$ & & $N \times 32$ & & activation$\coloneqq$ReLU \\
\textit{Linear} & & $N \times 32$ & & $N \times 64$ & & activation$\coloneqq$ReLU \\
\textit{Linear} & & $N \times 64$ & & $N \times 64$ & & activation$\coloneqq$ReLU \\
\textit{Linear} & & $N \times 64$ & & $N \times 32$ & & activation$\coloneqq$ReLU \\
\textit{Linear} & & $N \times 32$ & & $N \times 32$ & & activation$\coloneqq$ReLU \\
\textit{Linear} & & $N \times 32$ & & $N \times 1$ & & activation$\coloneqq$Identity \\
\bottomrule
\end{tabular}
}
\end{center}
\label{tab:parameters}
\end{table*}

\begin{table*}[h!]
\small
  \captionsetup{margin=0.3cm}
  \caption{\small\LTR~task \parencite{herman2021learn} results on \texttt{Track01} (Thruxton Circuit), for learning-\textit{free} agents, with respect to the task metrics: Episode Completion Percentage (\textbf{ECP}), Episode Duration (\textbf{ED}), Average Adjusted Track Speed (\textbf{AATS}), Average Displacement Error (\textbf{ADE}), Trajectory Admissibility (\textbf{TrA}), Trajectory Efficiency (\textbf{TrE}), and Movement Smoothness (\textbf{MS}). Arrows ($\uparrow\downarrow$) indicate directions of better performance, across agents. \textbf{Bold} results in tables \ref{tab:base_ltr_task_lf} and \ref{tab:base_ltr_task_lb} are generally best, however, asterisks (*)  indicate metrics which may be misleading, for incomplete racing episodes.}
  \label{tab:base_ltr_task_lf}
  \resizebox{\textwidth}{!}{
  \centering
  \begin{tabular}{lccccccc}
    \toprule
    \textbf{Agent} & \textbf{ECP} ($\uparrow$) & \textbf{ED}* ($\downarrow$) & \textbf{AATS} ($\uparrow$) & \textbf{ADE} ($\downarrow$) & \textbf{TrA} ($\uparrow$) & \textbf{TrE} ($\uparrow$) & \textbf{MS} ($\uparrow$) \\
    \midrule
    \texttt{HUMAN} & $\textbf{100.0}\pm 0.0$ & $78.6\pm 5.2$ & $\textbf{79.29}\pm 4.7$ & $2.4\pm 0.1$ & $0.93\pm 0.01$ & $\textbf{1.00}\pm 0.02$ & $\textbf{11.7}\pm 0.1$ \\ [1pt]
    \hline
    \texttt{Random} &  $0.50\pm 0.30$ & $\textbf{4.67}\pm 3.2$ & $11.90\pm 3.80$ & $1.5\pm 0.60$ & $0.81\pm 0.04$ & $0.33\pm 0.38^{*}$
    & $6.7\pm 1.1$ \\ [1pt]
    \texttt{MPC} &  $100.0\pm 0.0$ & $301.40\pm 10.10$ & $45.10\pm 0.0$ & $\textbf{0.90}\pm 0.10$ & $\textbf{0.98}\pm 0.01$ & $0.85\pm 0.03$ & $10.4\pm 0.60$ \\ [1pt]
    \bottomrule
  \end{tabular}
  }
\end{table*}
\begin{table*}[h]
\small
  \captionsetup{margin=0.3cm}
  \caption{\small\LTR~task \parencite{herman2021learn} results on \texttt{Track01} (Thruxton Circuit), for learning-\textit{based} agents.}
  \label{tab:base_ltr_task_lb}
  \resizebox{\textwidth}{!}{
  \centering
  \begin{tabular}{lccccccc}
    \toprule
    \textbf{Agent} & \textbf{ECP} ($\uparrow$) & \textbf{ED}* ($\downarrow$) & \textbf{AATS} ($\uparrow$) & \textbf{ADE} ($\downarrow$) & \textbf{TrA} ($\uparrow$) & \textbf{TrE} ($\uparrow$) & \textbf{MS} ($\uparrow$) \\
    \midrule
    \texttt{SAC} &  $61.61\pm 38.57$ & $272.75\pm256.51$ & $47.99\pm 30.9$  & $1.54\pm1.07$ & $\textbf{0.94}\pm0.02$ & $\textbf{0.28}\pm0.12$ & $11.84\pm2.12$ \\ [1pt]
    \texttt{SafeRandom} (ours), $\delta=3.0$ & $36.46\pm23.71$ & $654.37\pm447.05$ & $8.44\pm 1.37$ & $3.93\pm0.21$ & $0.81\pm0.10$ & $0.00\pm0.00$ & $13.21\pm1.88$ \\ [1pt]
    \texttt{SafeRandom} (ours), $\delta=4.2$ & $63.63\pm39.46$ & $761.80\pm494.65$ & $11.68\pm 1.07$ & $2.74\pm0.16$ & $0.90\pm0.07$ & $0.02\pm0.01$ & $\textbf{13.63}\pm2.01$ \\ [1pt]
    \texttt{SafeSAC} (ours), $\delta=3.0$ & $25.70\pm11.31$ & $66.90\pm23.22$ & $49.67\pm3.34$ & $1.35\pm0.05$ & $0.86\pm0.06$ & $0.14\pm0.05$ & $8.46\pm2.35$ \\ [1pt]
    \texttt{SafeSAC} (ours), $\delta=4.2$ & $49.05\pm41.66$ & $617.52\pm842.49$ & $33.83\pm26.21$ & $1.80\pm0.63$ & $0.91\pm0.12$ & $0.07\pm0.11$ & $10.03\pm2.75$ \\ [1pt]
    \texttt{\ourmethod} (ours) & $\textbf{79.94}\pm23.20$ & \textbf{59.19}$\pm29.99$ &\textbf{53.28}$\pm3.76$ & \textbf{0.99}$\pm0.17$ & $0.91\pm0.03$ & $0.22\pm0.03$ & $9.27\pm1.68$ \\ [1pt]
    \bottomrule
  \end{tabular}
  }
\end{table*}

\para{Implementation Details.}  For all experiments, we implemented the models using the \texttt{PyTorch} 1.8.0. We optimised both the performance and safety actor-critic with Adam  \cite{kingma2014adam}, with a learning rate of 0.003. We used $\gamma=0.99$ for the performance critic, and annealed $\gamma_S$ from 0.85 to 1 for the safety critic following \cite{fisac2019bridging}. We used $\tau=0.005$ for the performance critic, and $\tau=0.05$ for the safety critic. For both the performance and safety actor, we include the policy entropy term with $\alpha=0.2$. We used a batch size of 256, and a replay buffer size of 250,000.

\para{Computing hardware.} For rendering the simulator and performing local agent verification and analysis, we used a single GPU machine, with the following CPU specifications: Intel(R) Core(TM) i5-4690K CPU @ 3.50GHz; 1 CPU, 4 physical cores per CPU, total of 4 logical CPU units. The machine includes a single GeoForce GTX TITAN X GPU, with 12.2GB GPU memory. For generating multi-instance experimental results, we used a cluster of three multi-GPU machines with the following CPU specifications: 2x Intel(R) Xeon(R) Gold 5218R CPU @ 2.10GHz; 80 total CPU cores using a Cascade Lake architecture; memory of 512 GiB DDR4 3200 MHz, 16x32 GiB DIMMs. Each machine includes 8x NVIDIA GeForce RTX 2080 Ti GPUs, each with 11GB GDDR6 of GPU memory. Experiments were orchestrated on the these machines using Kubernetes, an open-source container deployment and management system.

All experiments were conducted using version 0.7.0.182276 of the Arrival Racing Simulator. The simulator and \LTR~framework \parencite{herman2021learn} are available for academic-use, here: \url{https://learn-to-race.org}.

\section{Additional Results}\label{sec:add_results}


\para{Performance of the SafeRandom agent.} Recall that the \texttt{SafeRandom} agent takes random actions and uses the safety value function precomputed from the nominal model. The optimal safety controller intervene whenever the safety value of the current state falls belong the safety margin. The safety margin is necessary because 1) the nominal model is a significant over-simplification of vehicle dynamics, and 2) the HJ Reachability computation does not take into consideration of the physical dimension of the vehicle. 

\begin{figure}[h!]
    \centering
    \includegraphics[width = \linewidth ]{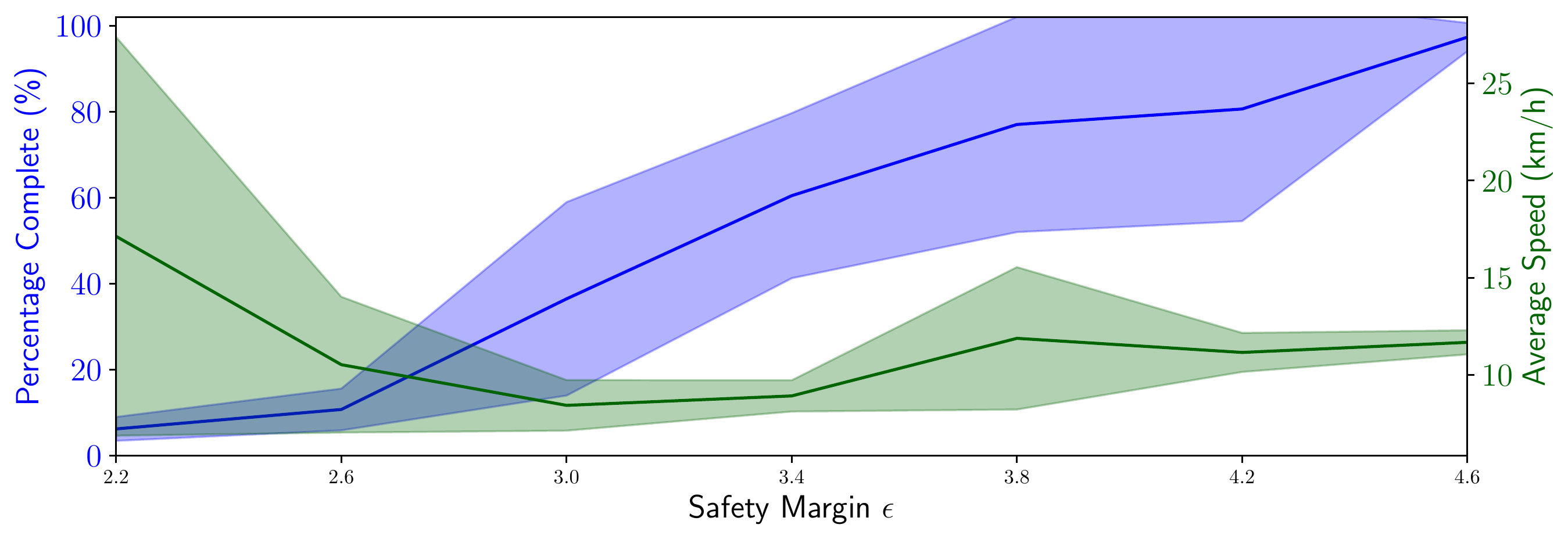}
    \caption{Performance of the \texttt{SafeRandom} agent at different safety margin (averaged over 10 random seeds)}
    \label{fig:SafeRandom}
\end{figure}

The performance of the \texttt{SafeRandom} agent at different safety margin is summarised in Figure \ref{fig:SafeRandom}. For safety margin $\epsilon\geq 4.2$, the \texttt{SafeRandom} agent can finish 80+\% of the lap, and thus we use $\epsilon=4.2$ as the safety margin for the \texttt{SafeSAC} agent. On the other hand, the performance decrease drastically when the safety margin is reduced to 3. 

\begin{figure}[h!]
    \centering
    \includegraphics[width = \linewidth ]{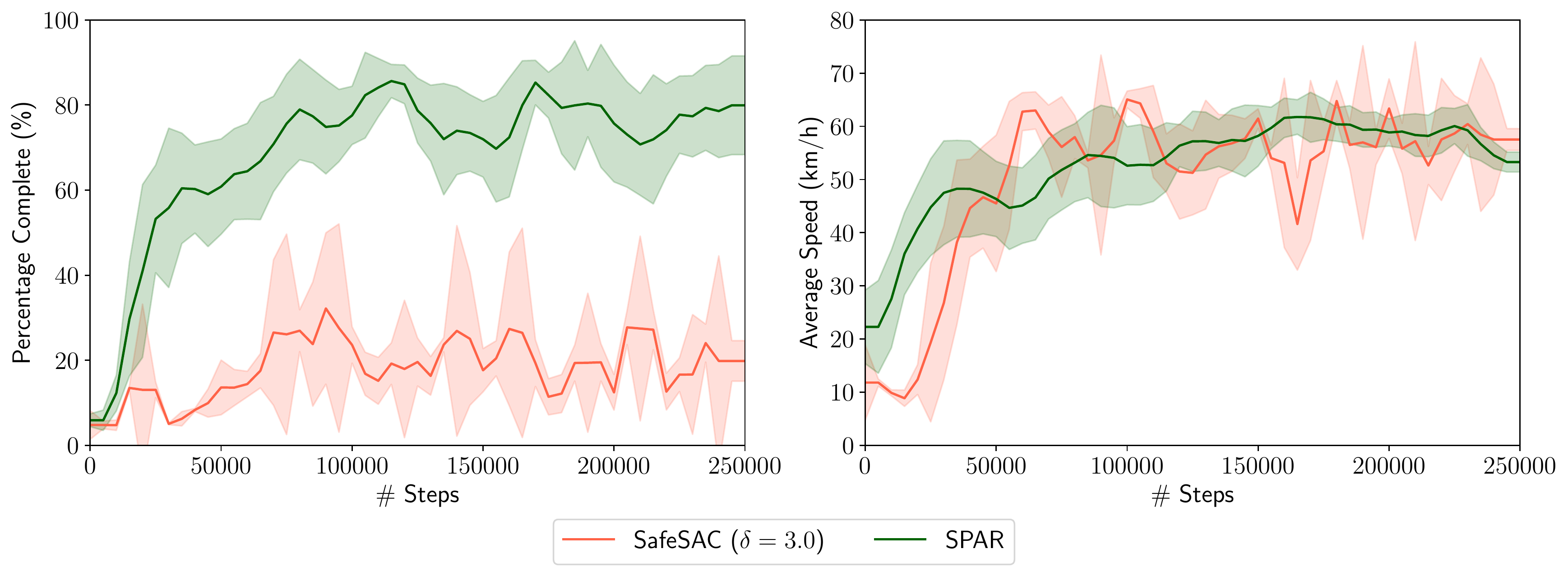}
    \caption{Performance of  \texttt{SafeSAC} ($\epsilon=3$) with comparison to \texttt{\ourmethod} }
    \label{fig:SafeSAC-3}
\end{figure}

\para{SafeSAC \& \ourmethod~performance with same safe margin.} While we choose the safety margin $\epsilon$ based on performance of the \texttt{SafeRandom} agent over a range of margins and our best engineering judgement, some may wonder if the superior performance of \ourmethod~over SafeSAC may be attributed to the use of different safety margins. Thus, we also show here the performance of a SafeSAC agent with the same safety margin as \ourmethod~in Figure \ref{fig:SafeSAC-3}. Given the smaller safety margin, the ECP is low initially, which is inline with the observation from \texttt{SafeRandom}. Furthermore, the ECP barely improves over time. As the performance agent learns to drive faster, it is increasingly difficulty for the static actor-critic to catch the vehicle in marginally safe states. 

\para{Learn-to-Race benchmark results.} In tables \ref{tab:base_ltr_task_lf} and \ref{tab:base_ltr_task_lb}, we follow \parencite{herman2021learn} in reporting on all of their driving quality metrics, for the \LTR~benchmark: Episode Completion Percentage (ECP), Episode Duration (ED), Average Adjusted Track Speed (AATS), Average Displacement Error (ADE), Trajectory Admissibility (TrA), Trajectory Efficiency (TrE), and Movement Smoothness (MS).

We highlight the fact that such metrics as TrA, TrE, and MS are most meaningful for agents that \textit{also} have high ECP results. Taking TrA, for example, safe policies score higher ECP values but may spend more time in inadmissible positions (as defined by the task, i.e., with at least one wheel touching the edge of the drivable area), compared to policies without a safety backup controller that may quickly terminate episodes by driving out-of-bounds (thus spending less time in the inadmissible positions). On the other hand, policies that have low completion percentages also have low ED scores, due to more frequent failures and subsequent environment resets.

We observe new state-of-the-art performance received by our approach, across the driving quality metrics, in the \LTR~benchmark.

\end{document}